\documentclass[times,twocolumn,final]{elsarticle}
\usepackage{framed,multirow}

\usepackage{amssymb}
\usepackage{latexsym}

\usepackage{url}
\usepackage{xcolor}
\usepackage{longtable}
\usepackage{float}
\usepackage{hyperref}

\definecolor{newcolor}{rgb}{.8,.349,.1}

\journal{Medical Image Analysis}

\begin{document}

\begin{frontmatter}

\title{Privacy Preserving Federated Learning in Medical Imaging with Uncertainty Estimation}%

\author[1,3]{Nikolas {Koutsoubis}\corref{cor1}}
\cortext[cor1]{Corresponding author: Nikolas Koutsoubis}
\cortext[cor1]{email: Niko.Koutsoubis@moffitt.org}
\author[3]{Yasin  {Yilmaz}}
\author[4]{Ravi P. {Ramachandran}}
% \fntext[fn1]{This is author footnote for second author.}
\author[2]{Matthew {Schabath}}
\author[1]{Ghulam {Rasool}}

%% Third author's email

\address[1]{Department of Machine Learning, Moffitt Cancer Center, Tampa, FL}
\address[2]{Department of Cancer Epidemiology, Moffitt Cancer Center, Tampa, FL}
\address[3]{Electrical Engineering Department, University of South Florida, FL,}
\address[4]{Electrical \& Computer Engineering Department, Rowan University, NJ}

% \finalform{10 May 2013}
% \accepted{13 May 2013}
% \availableonline{15 May 2013}
% \communicated{}

\begin{abstract}
Machine learning (ML) and Artificial Intelligence (AI) have fueled remarkable advancements, particularly in healthcare. Within medical imaging, ML models hold the promise of improving disease diagnoses, treatment planning, and post-treatment monitoring. Various computer vision tasks like image classification, object detection, and image segmentation are poised to become routine in clinical analysis. However, privacy concerns surrounding patient data hinder the assembly of large training datasets needed for developing and training accurate, robust, and generalizable models. Federated Learning (FL) emerges as a compelling solution, enabling organizations to collaborate on ML model training by sharing model training information (gradients) rather than data (e.g., medical images). FL's distributed learning framework facilitates inter-institutional collaboration while preserving patient privacy. However, FL, while robust in privacy preservation, faces several challenges. Sensitive information can still be gleaned from shared gradients that are passed on between organizations during model training. Additionally, in medical imaging, quantifying model confidence/uncertainty accurately is crucial due to the noise and artifacts present in the data. Uncertainty estimation in FL encounters unique hurdles due to data heterogeneity across organizations. This paper offers a comprehensive review of FL, privacy preservation, and uncertainty estimation, with a focus on medical imaging. Alongside a survey of current research, we identify gaps in the field and suggest future directions for FL research to enhance privacy and address noisy medical imaging data challenges.
\end{abstract}

\begin{keyword}
% %% or \MSC[2008] code \sep code (2000 is the default)
% \MSC 41A05\sep 41A10\sep 65D05\sep 65D17
%% Keywords
Federated Learning \sep Medical Imaging \sep Privacy Preservation \sep Uncertainty Estimation \sep Review
\end{keyword}

\end{frontmatter}

%\linenumbers

\section{Introduction}
Notably over the last decade, machine learning (ML) approaches have been leveraged for the analysis of medical imaging to improve the prediction of risk, early detection, diagnosis, treatment, and survival outcomes of numerous diseases \citep{Erickson2017,Latif2019,BarraganMontero2021,Willemink2020,pmid24892406,pmid26579733,pmid30720861,pmid32917666}. ML models have been used in clinical research applications leveraging radiological data, such as CT, MRI, PET, and more \citep{DAYARATHNA2024103046}. These ML models enable research scientists and clinical care teams to comprehend and interpret complex healthcare data accurately and efficiently. One key component in training effective ML models is curating large datasets necessary for training in the selected domain. This critical requirement presents a problem in the analysis of medical imaging due to privacy regulations regarding private health data, such as the Health Insurance Portability and Accountability Act (HIPAA)\citep{HIPAA1996} in the USA and the General Data Protection Regulation (GDPR)\citep{GDPR2016} in Europe. These regulations are designed to keep patient data secure and private, which makes it challenging to curate and combine large-scale training datasets across multiple sites. 

A conventional method of training ML models is centralized learning, which involves pooling data at a single location from all sources (e.g., sites). This may be challenging for medical datasets. A solution to circumvent centralized learning that has risen in popularity in recent years is Federated Learning (FL). FL was first proposed by Google for training ML models on edge devices without sharing client data \citep{pmlr-v54-mcmahan17a}. FL provides a method for training on data from multiple sites without data ever leaving the local site. This allows large-scale model training without violating privacy regulations that often hinder transferring and sharing data across sites. The overarching premise behind FL is that rather than transferring data, FL transfers training information (gradients) updates between sites. This permits multiple sites to act as clients and train a global model located at the central server. The global model is expected to outperform all local models on all data as it will learn from all local models without sharing private data.

In real-world settings, however, data distributions may differ across sites attributed to patient demographics, location, and other factors  \citep{Qu2021heter}. This violates the independent and identically distributed (\emph{i.i.d}) assumption of data and poses technical challenges for effective learning in a distributed FL setting. Many advances in FL techniques have come about in recent years, modifying the methods used to optimize FL's ability to learn from heterogeneous data while limiting communication costs to improve accuracy and efficiency. However, retaining client data locally is not enough to guarantee data privacy. It has been shown in multiple implementations of FL that through carefully planned methods, private data can be extracted from the communications that take place in an FL setting. Communications such as gradient updates can be used to reconstruct patient data through attacks such as gradient dis-aggregation \citep{Max2021Gdisag}, model inversion attack \citep{Wu2022invert}, and other methods \citep{Jere2021taxonomy}. Further enhancements in privacy preservation are critical for FL to be a paradigm-shifting technology to improve ML models in medical imaging. Methods such as differential privacy (DP) \citep{Dwork2006} and homomorphic encryption (HE) \citep{Gentry2009} have been leveraged to improve communications security. The advantages of these privacy preservation methods are that they can provide mathematical guarantees and show theoretical maximum accuracies based on different levels of privacy. An inherent trade-off exists in many privacy preservation techniques where, as privacy increases, model performance decreases. Sidestepping this trade-off is explored in further detail in Section \ref{sec:privacy}.

\begin{figure*}[ht]
\centering
\includegraphics[width=\textwidth]{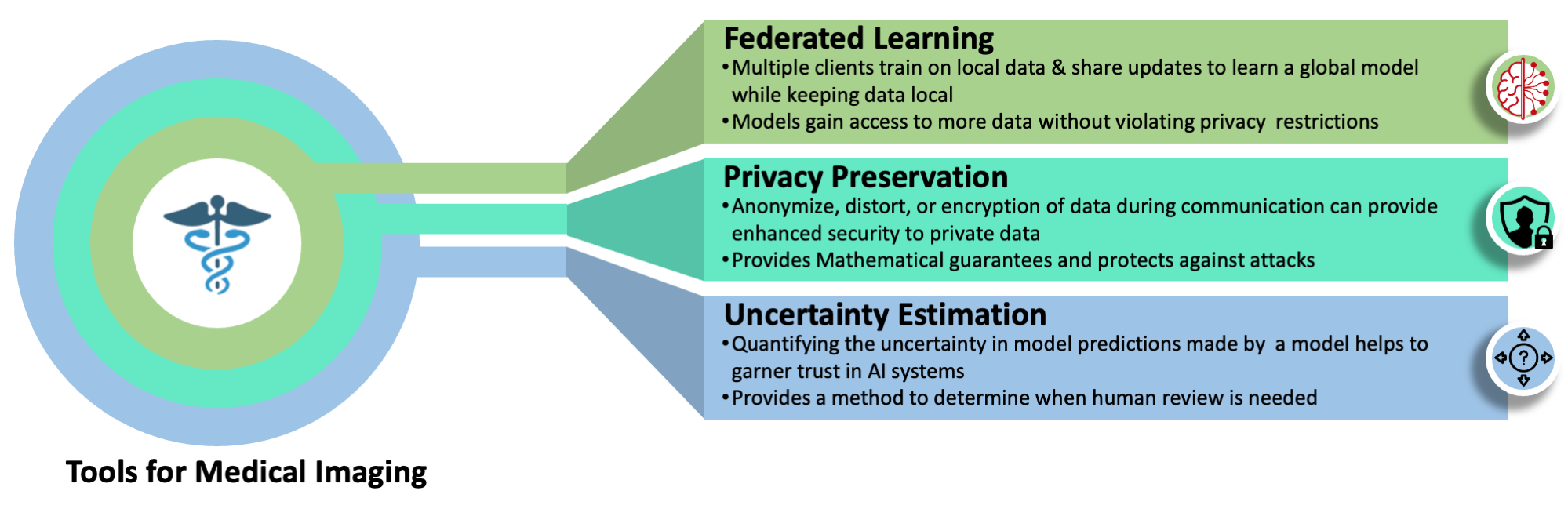} 
\caption{An overview of FL, privacy preservation, and uncertainty estimation is presented.}
\label{fig:1}
\end{figure*}

\begin{figure*}[ht]
\centering
\includegraphics[width=\textwidth]{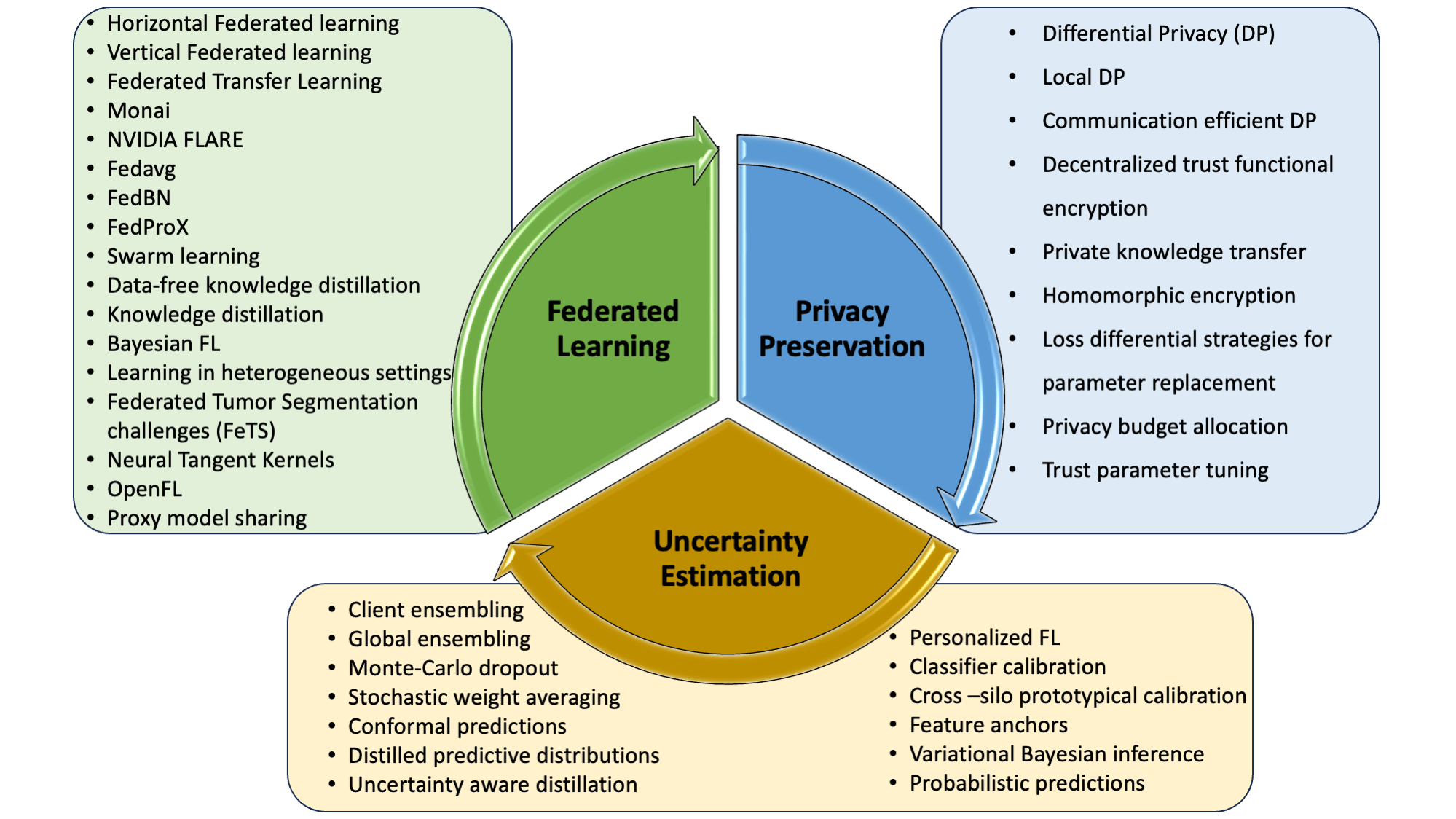}
\caption{Summary of topics covered in this review}
\label{fig:2}
\end{figure*}

Another key area in which FL needs to excel in medical imaging is uncertainty estimation, which is the process of measuring the reliability of a prediction/classification made by an ML model \cite{Linsner2021ApproachesTU}. ML models in the medical domain will ultimately be used to aid clinicians in the diagnosis and treatment of potentially life-threatening diseases. Hence, it is of utmost importance that the models have a way to notify users when they make uncertain or low-confidence predictions. Uncertainty estimation is a widely studied field \citep{Psaros2022}. However, FL presents unique challenges for uncertainty estimation. The non-i.i.d. nature of data in many FL applications, data imbalances on the client side, and variable computational overhead require modifications to traditional uncertainty estimation methods to work in FL environments successfully \citep{Linsner2021ApproachesTU}. Various methods have been explored in recent years and will be discussed further in Section \ref{sec:uncertain}.

FL holds the potential to significantly improve the role of ML models in the medical imaging domain, helping clinicians better diagnose and treat patients. However, FL alone cannot work in the medical imaging domain due to heterogeneous datasets, privacy regulations, and concerns about confidence in the model output. Extensive work has been conducted in recent years to mitigate and solve these challenges to elevate FL as a mainstream method for training ML models in medical imaging \citep{Darzidehkalani2022,kaissis2020secure,Mouhni2022}. Solving these issues would allow large-scale models to be trained on a wide variety of data, significantly improving the utility of these models for clinicians and researchers alike. This work presents a comprehensive review of the state-of-the-art methods of FL in the medical imaging domain. A summary of the aspects of FL and the topics covered in this paper can be seen in Figures \ref{fig:1} and \ref{fig:2}, respectively. The primary contributions of this work include:
\begin{itemize}
    \item A review of the current state-of-the-art (last 5 years) FL methods proposed in the medical imaging domain to deal with non-i.i.d. data in real-world settings.
    \item A review of the state-of-the-art privacy preservation methods extended to FL to guarantee data privacy.
    \item A review of uncertainty estimation methods effectively applied in FL settings, enabling trustworthy and reliable model development.
    \item Exploration of a number of real-world use cases of FL in the medical imaging domain and what can be learned from these success stories.
    \item Current challenges in FL, data privacy, uncertainty estimation, and potential opportunities and direction for future research.
\end{itemize}

The paper is organized as follows: Section \ref{sec:fl} presents a review of FL methods. Section \ref{sec:privacy} reviews the current state of privacy preservation in FL. Section \ref{sec:uncertain} explores recent advancements in uncertainty estimation in FL. Section \ref{sec:applications} covers the real-world applications of FL in medical imaging. Section \ref{sec:challenges} covers some current challenges and opportunities for research in FL for medical imaging. Finally, Section \ref{sec:conclusion} concludes the paper. A repository with links to papers reviewed in this work can be found in this  \href{https://github.com/Niko-k98/Awesome-list-Federated-Learning-Review/tree/main}{\textcolor{blue}{Awesome List}}.

\section{Federated Learning (FL)}
\label{sec:fl}
\begin{figure*}[ht]
\centering
\includegraphics[width=\textwidth]{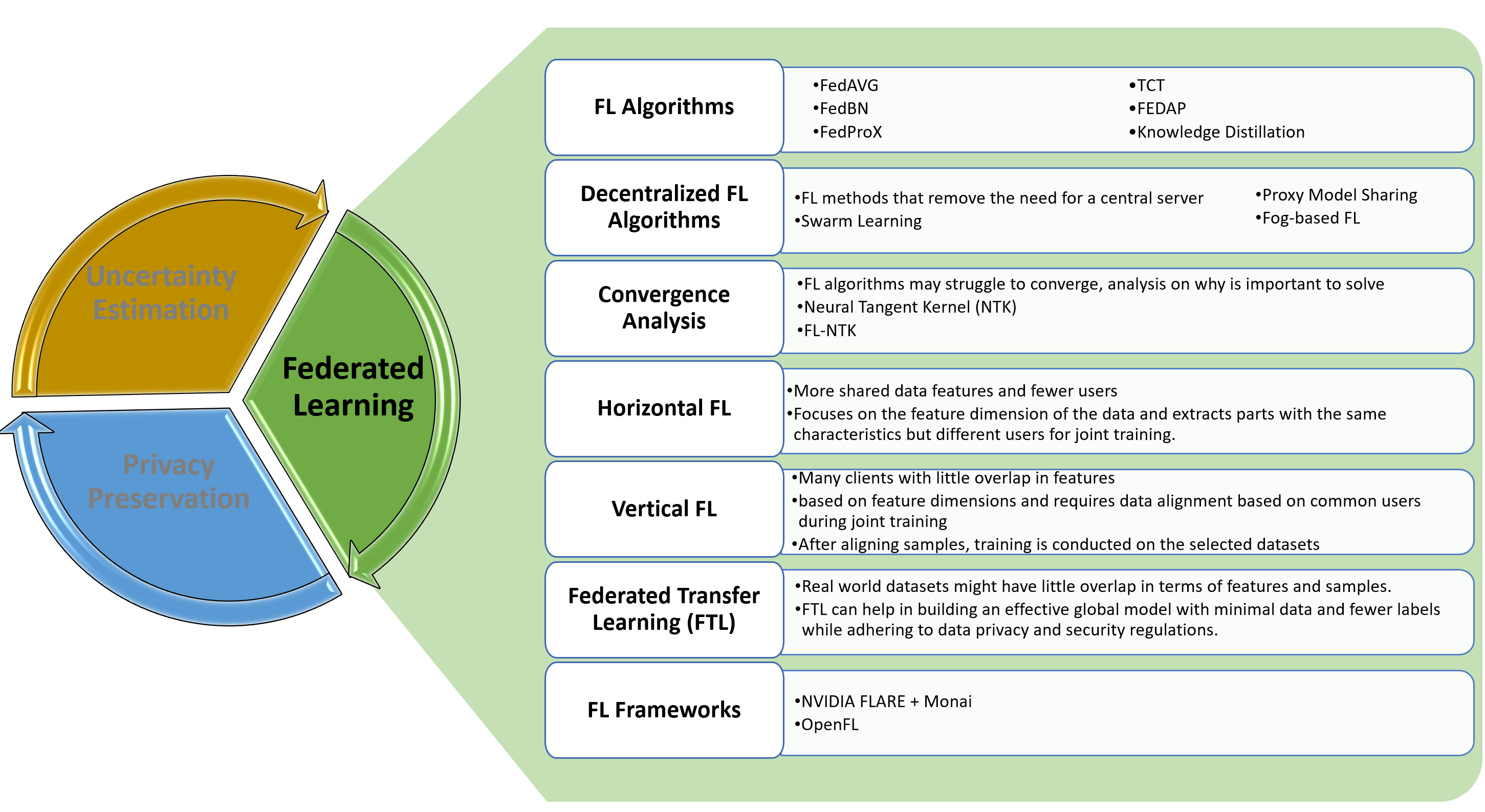}
\caption{Summary of FL topics}
\label{fig:3}
\end{figure*}

\begin{table*}[h!]
\centering
\caption{ FL algorithms}
\footnotesize{
\begin{tabular}{p{1cm} l p{1cm} p{1.2 cm} p{9cm}}
\hline \hline
\textbf{Algorithm} & \textbf{Ref}  & \textbf{Central Server} & \textbf{Local Forgetting} & \textbf{Summary} \\
\hline \hline 
FedAvg & \citep{pmlr-v54-mcmahan17a}   &yes & no & Train local models across various clients and then average the gradient updates at the central server to update the global mode; first proposed method of FL.\\
\hline 
FedProx & \citep{li2020federated}  &yes & no &  Excels in heterogeneous settings; generalization of the FedAvg algorithm; allows for partial updates to be sent to the server instead of simply dropping them from a federated round; adds proximal term that prevents any one client from having too much of an impact on the global model. \\
\hline
FedBN &\citep{Li2021FedBNFL} &yes & no & Addresses the issue of non-i.i.d. data by leveraging batch normalization; follows a similar procedure to Fed-Avg but assumes local models have batch norm layers and excludes their parameters from the averaging step. \\
\hline
TCT &\citep{Yu2022tct} & yes & yes & Train-Convexify-Train: Learn features with an off-the-shelf method (i.e., Fedavg) and then optimize a convexified problem obtained using the model’s empirical neural tangent kernel approximation; involves two stages where the first stage learns useful features from the data, and the second stage learns to use these features to generate a well-performing model.  \\
\hline
FedAP &\citep{lu2022personal} &yes & no & Learns similarities between clients by calculating distances between batch normalization layer statistics obtained from a pre-trained model;  these similarities are used to aggregate client models; each client preserves its batch normalization layers to maintain personalized features; the server aggregates client model parameters weighted by client similarities in a personalized manner to generate a unique final model for each client. \\
\hline
FedGen & \citep{zhu2021dfree} & yes & no & Learns a generator model on the server to ensemble user models' predictions, creating augmented samples that encapsulate consensual knowledge from user models; generate augmented samples that are shared with users to regularize local model training, leading to better accuracy and faster convergence. \\
\hline
FOLA & \citep{Liu2021ABF} &yes & yes & Bayesian federated learning framework utilizing online Laplace approximation to address local catastrophic forgetting and data heterogeneity; maximizes the posteriors of the server and clients simultaneously to reduce aggregation error and mitigate local forgetting. \\
\hline
FCCL &\citep{huang2022learn} &yes & yes & Federated cross-correlational and continual learning uses unlabeled public data to address heterogeneity across models and non-i.i.d data, enhancing model generalizability; constructs a cross-correlation matrix on model outputs to encourage class invariance and diversity; employs knowledge distillation, utilizing both the updated global model and the trained local model to balance inter-domain and intra-domain knowledge to mitigate local forgetting. \\
\hline
Swarm Learning &\citep{warnat2021swarm} & no & yes & Model parameters are shared via a swarm network, and the model is built independently on private data at the individual sites; only pre-authorized clients are allowed to execute transactions; on-boarding new clients can be done dynamically. \\
\hline
ProxyFL &\citep{Kalra2023Proxy} & yes & no & Clients maintain two models, a private model that is never shared and a publicly shared proxy model that is designed to preserve patient privacy; proxy models allow for efficient information exchange among clients without needing a centralized server; clients can have different model architectures. \\
\hline
FogML & \citep{butt2023fog} &no & no & Fog computing nodes reside on the local area networks of each site; fog nodes can pre-process data and aggregate updates from the locally trained models before transmitting, reducing data traffic over sending raw data. \\
\hline \hline
\end{tabular}
}
\label{table:fl_algorithms}
\end{table*}

FL was first proposed in \citep{pmlr-v54-mcmahan17a} with the FedAvg algorithm for training models on edge devices without exposing private data. This led to a paradigm shift in how ML models could be trained on sensitive and private data in distributed settings \citep{Wu2021,Mills2020}. The general idea of FL is that multiple clients train local ML models on their data and send gradient information to a central server to update a global model that, in theory, can outperform all local models on all data \citep{pmlr-v54-mcmahan17a}. FL is particularly powerful and applicable in medical ML research because the data never has to leave local sites, sidestepping many privacy regulations to protect private health data \citep{kaissis2020secure}. Additionally, medical imaging research often requires large volumes of data, frequently reaching terabytes or more, making data transfer challenging for centralized training. The first notable FL algorithm, FedAvg, trains local models across various clients and then averages the gradient updates at the central server to update the global model \citep{pmlr-v54-mcmahan17a}. The ML models trained with FL can perform at the same level of accuracy compared to those obtained using traditional centralized learning in many real-world medical applications \citep{sheller2020fed}. We have identified four main challenges/bottlenecks that still exist in creating a reliable FL framework for training medical imaging ML models \citep{wen2022survey,Linsner2021ApproachesTU}:
\begin{itemize}
    \item \textbf{Privacy and security challenges}: While data remains local to each site, private data can still be extracted from the gradient updates sent to and from the central server or between sites \citep{wen2022survey,Linsner2021ApproachesTU}. This calls for enhanced privacy preservation techniques, which will be discussed in Section \ref{sec:privacy}.
    
    \item \textbf{Heterogeneous and non-i.i.d. data distribution}: Due to variations in demographics, location, medical imaging equipment, and a variety of other factors, data between sites often violate the i.i.d. assumption, which can inhibit the models' ability to learn \citep{wen2022survey,Linsner2021ApproachesTU}. Various strategies are being developed to address this issue.
    
    \item \textbf{Significant communication overhead}: Since data never leaves the client's site, there must be significant communication between clients and the central server. This could be costly, especially with large gradient updates being sent very frequently. Communication-efficient FL frameworks must be designed for FL to succeed in medical imaging tasks \citep{wen2022survey,Linsner2021ApproachesTU}.
    
    \item \textbf{Uncertainty estimation}: The models used in medical imaging will aid clinicians in the diagnosis and treatment of potentially life-threatening diseases. Therefore, a method for quantifying the uncertainty in model prediction must be integrated into FL frameworks\citep{wen2022survey,Linsner2021ApproachesTU}. Due to the non-i.i.d. nature of the data in FL, new methods must be developed, or the existing methods must be adapted for FL. The current state of progress on the uncertainty estimation is discussed in Section \ref{sec:uncertain}.
\end{itemize}
A summary of the topics covered in this section can be seen in Figure \ref{fig:3}, and a table of comparisons as well as a short description of each  FL algorithm reviewed is presented in Table \ref{table:fl_algorithms}. 
\subsection{FL Algorithms}
 FL frameworks can be divided into three main categories \citep{wen2022survey}:
\begin{itemize}
    \item Horizontal FL - The dataset of each client has a larger overlap of data features than sites. This means there are more shared data features and fewer shared users. Horizontal FL focuses on the feature dimension of the data and extracts parts with the same characteristics but different users for joint training\citep{Liu2022TowardsMO}. Horizontal FL finds its usage in fields such as keyword spotting, emoji prediction, and blockchain \citep{Leroy2019Federated,Ramaswamy2019Federated,Fallah2020Personalized,pmlr-v54-mcmahan17a}. Horizontal FL offers significant benefits in terms of privacy and data security \citep{Liu2022TowardsMO}. It enables collaborative model training without exposing individual data points, thereby safeguarding sensitive information and enhancing privacy. By aggregating insights from diverse sources, horizontal FL improves model accuracy and robustness due to the abundance of data. This approach also reduces the risks associated with centralized data storage, such as breaches and misuse, and supports regulatory compliance efforts, like GDPR, by keeping data localized and within regulatory boundaries. Horizontal FL faces several challenges that can impact its efficiency and effectiveness \citep{Zhang2022Challenges}. The frequent exchange of model updates between participants and the central server, known as communication overhead, can be bandwidth-intensive. Additionally, heterogeneity in data distribution, device capabilities, and network connectivity can hinder model convergence and performance. As the number of participants grows, scalability issues arise, making it difficult to coordinate updates and maintain model quality. Moreover, Horizontal FL is susceptible to security threats, including model poisoning and inference attacks, which can compromise the model's integrity and potentially reveal sensitive information.
    
    \item Vertical FL - VFL is characterized by a scenario where client datasets have more overlapping users than overlapping data features. Each client dataset has more shared users, but data features are rarely duplicated. Vertical FL is based on feature dimensions and requires data alignment based on common users during joint training. After aligning samples from each participant's data, training is conducted on the selected datasets \citep{Liu2022VerticalFL}. Vertical FL has seen usage in the medical domain, financial domain, and malware detection \citep{Yang2023SurveyVFL,Khan2022CommunicationEfficientVFL,Serpanos2023MalwareDetectionVFL} Vertical FL proves to be exceptionally efficient in scenarios that demand the integration of datasets to uncover new insights, thereby facilitating cross-industry collaborations \citep{Yang2023SurveyVFL}. Additionally, its alignment with regulatory compliance mandates makes it an attractive option for industries looking to leverage collective data insights while maintaining strict privacy standards. The requirement for precise data alignment across different datasets introduces complexity, particularly with large-scale data from multiple sources, making the process challenging \citep{Yang2023SurveyVFL}. Vertical FL incurs significant communication overhead during model training, which can strain bandwidth and latency, thereby acting as a potential bottleneck. Its applicability is also somewhat limited, as it necessitates conditions where datasets share the same sample space but differ in feature space, restricting its use to specific scenarios. Moreover, despite its advantages in privacy preservation, Vertical FL remains vulnerable to security threats, including inference attacks \citep{FuLabelInferenceVFL}, where adversaries could potentially extract sensitive information from the model updates, thereby posing a risk to data privacy.
    \item Federated Transfer Learning - In many real-world scenarios, the datasets owned by each client can vary considerably. Federated transfer learning addresses these situations by enabling the construction of an effective global model despite minimal overlap in dataset features and samples. Federated transfer learning facilitates the development of models with limited data and fewer labels while adhering to data privacy and security regulations.
\end{itemize}
 
\subsubsection{FedProx} 
A variety of extensions to the original FedAvg algorithm have been proposed, such as FedProX \citep{li2020federated}. FedProX is an algorithm for FL that excels in heterogeneous data settings and serves as a generalization of the FedAvg algorithm, and FedAvg is considered a special case of FedProX \citep{li2020federated}. FedProX allows partial updates to be sent to the server instead of simply dropping them from a federated round and adding a proximal term that prevents any client from contributing too much to the global model, thereby increasing model stability. 
  
\subsubsection{FedBN} 
Another notable high-performing FL algorithm is FedBN \citep{Li2021FedBNFL}. This method outperforms both FedAvg and FedproX. This method addresses the issue of non-i.i.d. data by leveraging batch normalization. The authors introduce the concept of feature shift in FL as a novel category of a client’s non-i.i.d data distribution, where the following types of feature shifts can occur: 1) covariate shift: the marginal distributions $Pi(x)$ varies across clients, even if $Pi(y|x)$ is the same for all client; and 2) concept shift: the conditional distribution $Pi(x|y)$ varies across clients and $P(y)$ is the same, where features are $x$ and labels are $y$ on each client. FedBN uses the same premise as FedAvg, sending local updates and averaging at a coordinator. However, FedBN assumes local models have batch normalization layers and excludes their parameters from the averaging step. 

\subsubsection{Train-Convexify-Train}
Despite advancements offered by FedBN, Yu et al. point out the challenges due to the non-convexity of data \citep{Yu2022tct}. The authors point out that local models with different local optima can cause the global model to struggle to converge and hinder accuracy improvement \citep{Yu2022tct}. They find that the early layers of an FL model do learn useful features, but the final layers do not make use of them. That is, federated optimization applied to this non-convex problem distorts the learning of the final layers. To solve this issue, they propose a Train-Convexify-Train procedure, which involves learning features with an off-the-shelf method (i.e., Fedavg) and then optimizing a convexified problem obtained using the model's empirical neural tangent kernel (eNTK) approximation. This technique provided up to 37\% improvements in accuracy on dissimilar data over FedAvg alone. The convexity aspect attempts to compute a convex approximation of the model using its eNTK based on the concept of the neural tangent kernel (NTK) \citep{Jacot2018NeuralTK}. The eNTK approximates the fine-tuning of a pre-trained model. Train-Convexify-Train has two stages \citep{Yu2022tct}:
\begin{itemize}
    \item Stage 1 - Extract eNTK features from a trained FedAvg model. FedAvg is first used to train the model for a number of communication rounds. Then, each client locally computes sub-sampled eNTK features.
    \item Stage 2 - Decentralized linear regression with gradient correction. Given samples on each client $k$, first, normalize the eNTK inputs of all clients with a single communication round. Then, solve the linear regression problem with standard local learning rate and local steps $M$.
\end{itemize}
In this procedure, the first stage learns useful features from the data, and the second stage learns to use these features to generate a well-performing model \citep{Yu2022tct}.

\subsubsection{FedAP}  
Personalized FL involves training personalized models for various clients to deal with large data heterogeneity. Personalized FL balances the need for a generalized model and the benefits of localized, personalized models, making it a promising approach in applications where data privacy and customization are key concerns. One notable personalized FL algorithm is FedAP \citep{lu2022personal}. FedAP learns similarities between clients by calculating distances between batch normalization layer statistics obtained from a pre-trained model. These similarities are used to aggregate client models. Each client preserves its batch normalization layers to maintain personalized features. The server aggregates client model parameters weighted by client similarities in a personalized manner to generate a unique final model for each client. The authors evaluated FedAP on five healthcare datasets across modalities \citep{lu2022personal} including the public human activity recognition dataset, PAMAP2 \citep{Reiss2012ActivityMonitoring},  COVID-19 chest scan dataset \citep{Sait2020COVID19XRay}, MedMNIST, MedMNISTv2 \citep{yang2021medmnist,Yang2021MedMNISTv2}, the liver tumor segmentation benchmark \citep{Bilic2019LiTS}, and OrganAMNIST, OrganCMNIST, OrganSMNIST \citep{Xu2019OrganLocalization}. FedAP achieves more than $10\%$ accuracy over state-of-the-art FL models. FedAP converges faster than other methods, 10 rounds vs. more than 400 rounds for FedBN while being robust to varying hyperparameters and different degrees of non-i.i.d data distribution shifts among clients.

\subsubsection{FedGeN} 
Knowledge distillation is another emerging method for dealing with the challenge of data heterogeneity in FL. Based on work done by Hinton et al. \citep{hinton2015distilling}. Yang et al. \citep{Yang2024KD} implemented knowledge distillation for multi-organ segmentation in a federated paradigm. Knowledge distillation involves extracting useful knowledge from an ensemble of models. This has a natural extension to FL as multiple clients can serve as an ensemble from which knowledge can be distilled. One 
successful implementation of Federated knowledge distillation is FedGen \citep{zhu2021dfree}. Most knowledge distillation methods require a proxy dataset to distill the knowledge. FedGen proposes a data-free method, thereby removing the need for this proxy and making knowledge distillation more accessible to applications where a proxy dataset cannot be created due to lack of data or privacy restrictions \citep{zhu2021dfree}. FedGen learns a generator model on the server that ensembles the prediction rules from user models. This generator can produce augmented samples that convey consensual knowledge from the user models. The generator is shared with the users and provides additional samples that regularize local model training. This distills the aggregated knowledge into the user models. Sharing the lightweight generator introduces minimal communication overhead and increases security because the full model is not shared. FedGen achieves better accuracy and faster convergence than state-of-the-art methods under heterogeneous settings. Benefits are especially notable as the data heterogeneity increases. To model non-iid data distributions, the authors follow the work done by Lin et al. \citep{Lin2020EnsembleDistillation} and Hsu et al. \citep{Hsu2019NonIdenticalDistribution}, using a Dirichlet distribution Dir $(\alpha)$ in which a smaller $\alpha$ indicates higher data heterogeneity, as it makes the distribution more biased for a user $k$. 

\subsubsection{Federated Online Laplace Approximation (FOLA)}
In addition to data heterogeneity, another challenging issue in FL is local catastrophic forgetting. That is, the local models forget the specific attributes of their data when they are updated with the global model because the global weights overwrite the model weights. Catastrophic forgetting is also a challenge in continual learning \citep{khan2024brain, Khan223Importance}. To combat this problem and data heterogeneity, a Bayesian FL algorithm with online Laplace approximation is proposed by \citep{Liu2021ABF}. Federated Online Laplace Approximation (FOLA) operates by integrating Bayesian principles with an online approximation approach, thereby effectively estimating probabilistic parameters of both global and local models in real-time. This approach addresses aggregation error and local forgetting by efficiently approximating Gaussian posterior distributions in a distributed setting. A Gaussian product method is used to construct a global posterior on the server side and a prior iteration strategy to update the local posteriors on the client sides, both of which are easy to optimize. Successfully maximizing these posteriors of the server and clients simultaneously reduces aggregation error and local forgetting \citep{Liu2021ABF}.

\subsubsection{Federated Cross-Correlation and Continual Learning (FCCL)}
Another method that addresses local forgetting is Federated Cross-Correlation and Continual Learning (FCCL) \citep{huang2022learn}. To handle heterogeneity across models and non-i.i.d data, FCCL leverages unlabeled public data and constructs a cross-correlation matrix on the models' logit outputs. This encourages correlation among the same logit dimensions (class invariance) while de-correlating different dimensions (class diversity) to learn a more generalizable representation. To alleviate catastrophic forgetting during local updates, FCCL employs knowledge distillation using the updated global model (to retain inter-domain information learned from others) and a trained local model (to retain intra-domain information) without leaking privacy. This continually balances knowledge, helping to alleviate catastrophic forgetting. 

\subsection{Decentralized FL algorithms}
Many FL algorithms utilize a central server where local model updates are sent from clients to update the local model. However, many FL implementations do not utilize a central server \citep{warnat2021swarm,Kalra2023Proxy, butt2023fog}. This subsection describes some recent decentralized FL algorithms. 

\subsubsection{Swarm Learning}
Swarm learning \citep{warnat2021swarm} is a decentralized learning method that unites edge computing with blockchain-based peer-to-peer networking, eliminating the need for a central coordinating server. This approach combines decentralized hardware infrastructures and distributed ML using standardized engines with a permissioned blockchain to securely onboard members, dynamically elect the leader, and merge model parameters. Model parameters are shared via a swarm network, allowing independent model building on private data at individual sites. Security and confidentiality are ensured by the permissioned blockchain (making each client well-defined as a participant), which restricts execution to pre-authorized clients. New clients can be dynamically onboarded.

\subsubsection{ProxyFL}
ProxyFL \citep{Kalra2023Proxy} was proposed by Kalra et al. as a communication-efficient scheme for decentralized FL. Each client maintains two models: a private model that is never shared and a publicly shared proxy model that is designed to preserve privacy. Using proxy models allows for efficient information exchange among clients without needing a centralized server. One massive step forward from traditional FL is that ProxyFL allows each client to have model heterogeneity, meaning that each client's private model can have any architecture. The proxy models also utilize DP to improve privacy. ProxyFL can outperform existing alternatives with much less communication overhead and stronger privacy. 

\subsubsection{Fog-FL}
Fog-FL enhances the efficiency and reliability of computing using a decentralized computing infrastructure between the data source and the cloud, known as fog computing\citep{butt2023fog}. It provides computing, storage, and networking services closer to where the data is generated, i.e., at the network edge. Fog computing nodes reside on the local area networks of each medical institution. These Fog computing nodes can pre-process data and aggregate updates from the locally trained models before transmitting them to the central FL server in the cloud. Fog computing reduces the data traffic and latency compared to sending all raw data directly to the cloud. It also enhances privacy and security as sensitive data stays on the local network.

\subsection{Software Frameworks for FL Implementation}
An FL project involves many moving parts, requiring coordination between all clients and the central server. To facilitate this process, several open-source frameworks have been developed to aid in setting up and managing federated runs. 

\subsubsection{OpenFL} 
OpenFL is an open-source Python library for FL\citep{Foley_2022}. This framework supports both Tensorflow and PyTorch projects, and the workflow is defined by a federation plan that all sites agree upon before beginning. OpenFL uses a static network topology with clients connecting to a central aggregating server over encrypted channels. OpenFL was designed with medical imaging in mind and is set up for horizontal FL but can be extended to other types. OpenFL allows easy migration of centralized ML models into a federated training pipeline and is designed for real-world scalability and trusted execution.  

\subsection{NVIDIA MONAI, FLARE, and Clara} 
NVIDIA MONAI, FLARE, and Clara are three integral frameworks developed by NVIDIA to advance FL and medical imaging \citep{MONAI-Deploy}, \citep{NVIDIA-CLARA}, \citep{Roth2022FLARE}. Medical Open Network for AI (MONAI) is an open-source framework optimized for healthcare, providing domain-specific tools and deep learning models to streamline the development of medical imaging solutions. It integrates seamlessly with Federated Learning Application Runtime Environment (FLARE), another open-source SDK by NVIDIA designed to facilitate FL. FLARE supports various FL architectures and incorporates robust privacy-preserving techniques like DP and HE. Clara, specifically Clara Train, is a medical imaging platform that leverages FLARE to enable FL within its ecosystem.  Some key components of this NVIDIA FLARE include  \citep{Roth2022FLARE}:
\begin{itemize}
    \item An FL simulator for rapid development and prototyping.
    \item A dashboard for simplified project management, secure provisioning, and deployment orchestration.
    \item Reference FL algorithms like Fedavg, fedproX, and FedOpt, with workflows like scatter and gather.
    \item Privacy preservation options like DP, HE, and others.
    \item Specification-based API for custom implementations
    \item Tight integration with frameworks like MONAI.
\end{itemize}

\subsection{Convergence of Model Learning in FL}
ML models trained using federated runs can struggle to converge due to the non-i.i.d nature of the model training data. Conventional ML training convergence analysis methods are not necessarily suited for FL settings. Huang et al. propose FL Neural Tangent Kernel (FL-NTK) \citep{huang2021ntk} to perform convergence analysis of FL algorithms. FL-NTK analyzes the convergence and generalization of FL algorithms in the context of over-parameterized Rectified Linear Unit (ReLU) neural networks. The authors show that FL-NTK converges to a global-optimal solution at a linear rate with properly tuned learning parameters, such as quartic width \citep{huang2021ntk}. The framework offers insights into different FL optimization and aggregation methods. The authors conducted experiments using the CIFAR-10 \citep{cifar10} dataset and ResNet56 model and explored FL with different numbers of clients ($N$) and two types of data distributions, i.i.d and non-i.i.d \citep{huang2021ntk}. \\

\section{Privacy Preservation in FL}
\label{sec:privacy}

\begin{figure*}[ht]
\centering
\includegraphics[width=\textwidth]{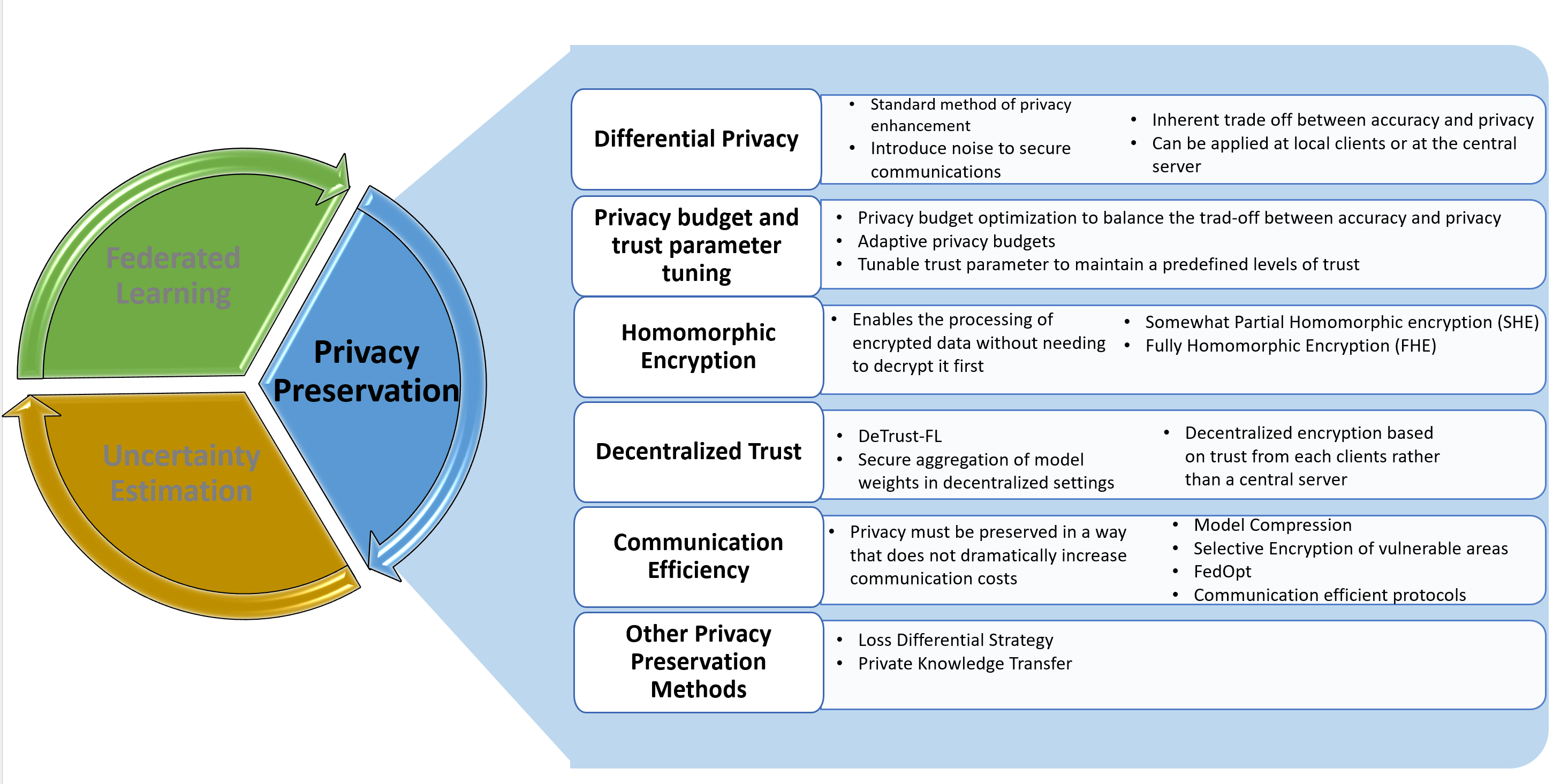} 
\caption{Summary of privacy preservation methods in FL.}
\label{fig:4}
\end{figure*}

\begin{table*}[htpb]
\centering
\caption{Privacy Preservation Methods in FL.}
\footnotesize{
\begin{tabular}{p{1.5cm} l p{1.5cm} p{2.0cm}  p{9cm}}
\hline \hline
\textbf{Algorithm} & \textbf{Ref}  & \textbf{Differential Privacy (DP)} & \textbf{Homomorphic Encryption (HE)} & \textbf{Summary} \\
\hline \hline
NbAFL & \citep{9069945}   &yes & no &  Noising before aggregation FL (NbAFL) Uses K-random scheduling to optimize the privacy and accuracy trade-off by introducing artificial noise into the parameters of each client before aggregation. \\
\hline
Adaptive privacy & \citep{nampalle2023vis}  &yes & no & Adaptive allocation of the privacy budget across FL iterations; the aim is to optimize the use of the privacy budget based on the data distribution and the model's learning status; higher privacy budget allocated earlier in training, and lower budget later to optimize privacy budget utilization. \\
\hline
FedOpt &\citep{asad2020opt} &yes & yes &  Utilizes sparse compression and HE for secure gradient aggregation and DP for enhanced privacy.\\
\hline
SHEFL & \citep{TRUHN2023103059} & yes & yes &  Somewhat homomorphically encrypted FL (SHEFL); only communicating encrypted weights; all model updates are conducted in an encrypted space.   \\
\hline
Hybrid Approach &\citep{truex2019hybrid} &yes & yes &  Combining DP with secure multiparty computation enables this method to reduce the growth of noise injection as the number of parties increases without sacrificing privacy; the trust parameter allows for maintaining a set level of trust.\\
\hline
PrivateKT & \citep{Qi2023} & yes & no & Private knowledge transfer method that uses a small subset of public data to transfer knowledge with local DP guarantee; selects public data points based on informativeness rather than randomly to maximize the knowledge quality. \\
\hline
Multi-RoundSecAgg & \citep{So2023secure} &yes & no & Provides privacy guarantees over multiple training rounds; develops a structured user section strategy that guarantees the long-term privacy of each use. \\
\hline
LDS-FL &\citep{wang2023LDS}  &no & no & Maintain the performance of a private model preserved through parameter replacement with multi-user participation to reduce the efficiency of privacy attacks. \\
\hline
DeTrust-FL & \citep{Xu2022DeTrustFLPF}   & no & no &  Provides secure aggregation of model updates in a decentralized trust setting; implements a decentralized functional encryption scheme where clients collaboratively generate decryption key fragments based on an agreed participation matrix. \\

\hline \hline
\end{tabular}
}
\label{table:Privacy_Presevation}
\end{table*}

Ensuring the secure processing of protected and identifiable information is a critical priority in the medical field. Federal regulations strictly prohibit the sharing of patient data to prevent privacy breaches. FL addresses this issue by keeping data localized at each site. However, even with data remaining local, privacy can still be compromised. The gradient updates exchanged between clients and the server can potentially reveal information about the training data, leading to privacy leaks. The attacks like the gradient dis-aggregation attack \citep{Max2021Gdisag} highlight the need for enhanced privacy measures in FL to safeguard sensitive information effectively. The topics covered in this section are summarized in Figure \ref{fig:4} and Table \ref{table:Privacy_Presevation}.

\subsection{Differential Privacy (DP)}
One of the most popular methods for privacy preservation is DP \citep{Dwork2006}, which introduces noise into the gradients to prevent private information leakage. DP has been used in medical imaging applications \citep{LU2022102298}. The DP method provides mathematical guarantees of privacy. However, the guarantees come at the cost of accuracy and convergence \citep{li2019privacy,nanayakkara2022visualizing,xu2021privacy}. 

\subsubsection{Noising before model aggregation FL (nbAFL)}
nbAFL adds artificial noise to parameters at the client side before aggregation to ensure DP and then proposes K-random scheduling to find optimal convergence \citep{9069945}. In K-random scheduling, K clients are chosen randomly to participate in the aggregation process, ensuring not all information is communicated in every round. This makes it harder for attackers to extract useful information. An optimal value of $K$ must be found for a given privacy level. This trade-off is often referred to as privacy budget allocation. nbAFL can balance the privacy level and the desired accuracy based on the application.

\subsubsection{Adaptive privacy budget allocation}
One method for a well-designed privacy budget was proposed by Nampelle et al. \citep{nampalle2023vis}. The authors demonstrate that strategic calibration of the privacy budget in DP can uphold robust performance while providing substantial privacy guarantees. They propose an adaptive privacy budget allocation strategy for FL rounds that best updates the privacy budget in each round based on the data distribution and model learning progress. The key aspect of their methods is the adaptive allocation of the privacy budget across FL iterations. The aim is to optimize the use of the privacy budget based on the data distribution and the model's learning status. The strategy is to allocate more of the budget in the earlier iterations of FL, where the models learn more from the data. Later iterations have less privacy budget allocated as the gradients have less information about the data. This design optimizes the trade-off between learning and privacy.

\subsubsection{Privacy-performance trade-offs} Differentially private FL can provide comparable performance to centralized learning \citep{Adnan2021}.  The authors in \citep{Adnan2021}implement a two-step method for DP. First, multiple patches are extracted, and a mosaic is formed for training using a memory network and an attention-based multiple instant learning algorithm that provides privacy bounds locally. The local models with DP are then aggregated at the central server. The method was tested on simulated real-world data in both i.i.d and non-i.i.d. settings. Contrarily, Choudhury et al. \citep{Choudhury2019DifferentialPF} found that although DP guarantees a given level of privacy as set by its parameters, it significantly deteriorates the utility of the FL model. The model's performance can only be preserved with a very large number of sites, on the order of $10^3$, but suffers severely in cases with fewer sites. Such an assumption of large-scale setup is unrealistic for healthcare applications, where sites are typically hospitals or providers, and each site may not have sufficient data for independently training deep learning models.

\subsection{Homomorphic Encryption (HE)} 
While DP has proven useful, HE has also been extensively explored in FL \citep{Dhiman2023,Stripelis2021,Burlachenko2023}. HE is a form of encryption that allows computations (mathematical operations) to be carried out on ciphertexts, generating encrypted results that, when decrypted, match the result of operations performed on the plain data. Thus, the data can be encrypted and shared with a third party for processing without the third party having access to the decrypted data.

\subsubsection{FedOpt} 
FedOpt \citep{asad2020opt} provides a communication-efficient method for privacy preservation in FL. This method uses a novel sparse compression algorithm to reduce communication overhead by extending top-k gradient compression with a downstream compression mechanism. The authors adopt lightweight HE for efficient and secure aggregation of gradients, using additive HE without key-switching to increase plain-text space \citep{asad2020opt}. The authors also employ DP. FedOpt is robust to user dropouts during training, with little impact on accuracy. Evaluations show that FedOpt outperforms state-of-the-art approaches like FedAvg and PPDL in model accuracy, communication efficiency, and computation overhead. 

\subsubsection{Somewhat Homomorphic Encryption (SHE)}
SHE is a subset method of HE \citep{acar2017survey} and is a type of encryption that allows for a limited number of arithmetic operations on encrypted data. Unlike Fully Homomorphic Encryption (FHE), which supports unlimited operations, SHE has constraints on the number and type of operations that can be executed. SHE is generally more efficient than FHE because it deals with a restricted set of operations \citep{acar2017survey}. This makes it more practical for applications where the computational overhead of FHE would be prohibitive \citep{acar2017survey}. 

\subsubsection{Somewhat Homomorphically Encrypted FL (SHEFL)}
Truhn et al. leveraged SHE and proposed SHEFL, which enables multiple parties to co-train ML models for pathology and radiology images securely, reaching state-of-the-art performance with privacy guarantees while requiring negligible extra computational resources \citep{TRUHN2023103059}. SHEFL provides a solution to privacy concerns by only communicating encrypted weights, and model updates are conducted in an encrypted space. The authors implement SHEFL on two clinical use cases - segmenting brain tumors \citep{menze2014multimodal} and predicting biomarkers from histopathology slides in colorectal cancer\citep{TCGACOADREAD}. The models trained with SHEFL are on par with regular FL while providing privacy guarantees, showing that encryption does not negatively impact accuracy \citep{TRUHN2023103059}. The methods only encrypt the vulnerable areas of the FL with a less than $5\%$ increase in train time or compute. The authors show the encryption/decryption process is negligible compared to backpropagation. When faced with an inversion attack, a normal FL algorithm could have its data reconstructed in 120 iterations, but with SHELF, the data was secure after 40,000 iterations  \citep{TRUHN2023103059}.

\subsection{Other methods of privacy preservation}
In addition to DP and HE, other methods have been used to preserve privacy in FL, such as using the aforementioned methods in conjunction with other techniques to optimize security. 

\subsubsection{A hybrid approach} 
A hybrid approach to privacy-preserving FL is proposed in \citep{truex2019hybrid} that uses DP and secure multi-party computation to balance the trade-off between privacy and accuracy. Combining DP with secure multiparty computation enables this method to reduce the growth of noise injection as the number of parties increases without sacrificing privacy while maintaining a pre-defined rate of trust with a tuneable trust parameter that can account for various scenarios. The trust parameter $t$ refers to the minimum number of honest, non-colluding parties the system assumes \citep{truex2019hybrid}. This parameter captures the degree of possible adversarial knowledge by specifying the maximum number of colluding parties the system can tolerate while still providing privacy guarantees. The noise added by each honest party depends on $t$. As $t$ decreases (less trust), more noise needs to be added by each honest party to account for more potential colluders. The threshold encryption scheme uses $t$ to set the threshold so that no set of parties less than this threshold can decrypt data. This prevents smaller colluding groups from learning honest parties' data. The trust parameter $t$ is useful in preventing dishonest parties from acting as clients and gaining access to honest clients' data. 

\subsubsection{PrivateKT} 
One method that leverages DP to implement private knowledge transfer is PrivateKT \citep{Qi2023}, a private knowledge transfer method that uses a small subset of public data to transfer knowledge with local DP guarantees. This method actively selects public data points based on the information contents rather than randomly to maximize the knowledge quality. The knowledge transfer method contains three steps:
\begin{itemize}
    \item \emph{Knowledge Extraction}: The clients train their models on local private data, then make predictions with their models on a small set of specifically selected public data points (KT data) \citep{Qi2023}. This process extracts knowledge from private data and uses it to make predictions about public data.
    \item  \emph{Knowledge Exchange}: The clients locally add DP to the public data predictions using a randomized response mechanism to guarantee DP. These DP predictions are then sent to the central server \citep{Qi2023}.
    \item \emph{Knowledge Aggregation}: The central server aggregates DP predictions from all clients and stores them in a knowledge buffer \citep{Qi2023}. 
\end{itemize}

knowledge transfer can securely transfer data between models and also provide uncertainty estimation through its functionality. Two methods are implemented to improve the effectiveness of knowledge transfer on a small amount of public data.
\begin{itemize}
    \item \emph{Importance Sampling} - The model's uncertainty is measured on each public data point using information entropy, and a higher sampling probability is assigned to data with higher uncertainty \citep{Qi2023}. This maximizes the information and quality of the knowledge sampled in a small dataset \citep{Qi2023}.
    
    \item \emph{Knowledge Buffer} - The server stores the DP aggregated predictions from clients in a buffer that maintains a history of past aggregated knowledge  \citep{Qi2023}. This buffer is used to fine-tune the global model, encoding historical knowledge to help mitigate the limitations of a small dataset  \citep{Qi2023}. A knowledge buffer is typically implemented during the knowledge aggregation step.
\end{itemize}

PrivateKT is tested on MNIST \citep{lecun1998gradient}, METText, and a Kaggle X-ray image dataset for pneumonia detection. Under a strict privacy budget, PrivateKT reduces the performance gap with centralized learning by up to $84\%$ compared to other FL methods \citep{Qi2023}. 

\subsubsection{Multi-RoundSecAgg} 
So et al. point out that many privacy preservation methods only provide privacy guarantees for a single communication round \citep{So2023secure}. The authors propose Multi-RoundSecAgg, which provides privacy guarantees over multiple training rounds. The authors also introduce a new metric to quantify the privacy guarantees of FL over multiple training rounds and develop a structured user section strategy that guarantees the long-term privacy of each user. Multi-RoundSecAgg contains two components: (1) a family of sets of users that satisfy the multi-round privacy requirement,  and (2) a set from this family to satisfy a fairness guarantee. The authors found a trade-off between long-term privacy guarantees and the number of participants. As the average number of users increases, long-term privacy becomes weaker \citep{So2023secure}. Random user selection schemes are shown to provide very weak multi-round privacy. After sufficient rounds (linear in number of users), the server can reconstruct all user models. Multi-RoundSecAgg is a structured user selection strategy with provable multi-round privacy. It partitions users into batches that always participate together. Multi-RoundSecAgg provides a trade-off between privacy and convergence rate. More privacy reduces the average number of users per round, slowing down training. The authors show that structured user selection is necessary for long-term privacy \citep{So2023secure}.

\subsubsection{Loss Differential Strategy for Parameter replacement (LDS-FL)}  
One method for privacy preservation that takes a different approach entirely is the loss differential strategy for parameter replacement (LDS-FL) \citep{wang2023LDS}. The key idea of this strategy is to maintain the performance of a \emph{private model} preserved through parameter replacement with multi-user participation. LDS-FL introduces a public participant that shares parameters to enable other private participants to construct \emph{loss differential models} without exposing private data. These satisfy an inequality that bounds loss on public, private, and other data. Wang et al. propose a loss differential strategy (LDS) where private participants replace some public parameters with their own to create models that resist privacy attacks. This balances privacy and accuracy \citep{wang2023LDS}. The authors formally prove the privacy guarantees of the LDS approach against membership inference attacks. Experiments show that LDS-FL reduces attack accuracy while maintaining model accuracy. The multi-round LDS algorithm enables participants to iteratively construct loss differential models in a privacy-preserving and convergent way during FL \citep{wang2023LDS}. Comprehensive experiments on image datasets demonstrate that LDS-FL reduces attack accuracy by over $10\%$ on MNIST while reducing model accuracy by just $0.17\%$ \citep{wang2023LDS}. LDS-FL outperforms DP defenses in accuracy and attack resistance. This method does not use DP or HE but rather provides an alternative method for preserving privacy, suggesting other ways to solve the issue of privacy preservation in FL. 

\subsubsection{DeTrust-FL} 
DeTrust-FL \citep{Xu2022DeTrustFLPF} proposes a solution to enhance the privacy of FL in a decentralized setting and provides secure aggregation of model updates in a decentralized trust setting. DeTrust-FL improves other PPFL methods by not relying on a centralized trusted authority and vulnerability to inference attacks like dis-aggregation attacks. DeTrust uses a decentralized functional encryption scheme where clients collaboratively generate decryption key fragments based on an agreed participation matrix. using a participation matrix provides transparency and control over the aggregation process, as all participants know what they agree to. Detrust-FL incorporates batch partitioning to prevent dis-aggregation attacks and encrypts model updates with round labels to prevent replay attacks. The authors show that DeTrust-FL achieves state-of-the-art communication efficiency and reduces reliance on a centralized trust entity \citep{Xu2022DeTrustFLPF}.

\subsection{Privacy preserving FL Frameworks}
This section presents Frameworks that have been created to streamline the process of privacy preservation in FL.

\subsubsection{Argonne Privacy-Preserving Framework (APPFL)} 
APPFL provides an open-source Python package that provides tools for users to run FL experiments with additional privacy preservation tools \citep{Ryu2022appfl}. There are five main components of this framework:
\begin{itemize}
    \item FL algorithms
    \item DP schemes
    \item Communication protocols
    \item Neural network models
    \item Data for training and testing
\end{itemize}

 The APPFL framework provides users with the tools to conduct their experiments with FL and allows for flexibility in model choice and the ability to implement custom models \citep{Ryu2022appfl}. APPFL also provides a communication-efficient inexact alternating direction method of multipliers (IIADMM) based on the Alternating Direction Method of Multipliers (ADMM) \citep{boyd2011distributed}. The  IIADMM algorithm significantly reduces the amount of information transferred between the server and the clients compared to similar algorithms.

\subsubsection{Privacy-preserving Medical Image Analysis (PriMIA)} 
PriMIA is an open-source software framework for differentially private, securely aggregated FL and encrypted inference on medical imaging data \citep{KaissisEnd}. The authors tested PriMIA using a real-life case study on pediatric chest X-rays. They found their privacy-preserving federated model was on par with local non-securely trained models. They theoretically and empirically evaluate the framework’s performance and privacy guarantees and demonstrate that the protections provided prevent the reconstruction of usable data by a gradient-based model inversion attack \citep{KaissisEnd}. The authors successfully employ the trained model in an end-to-end encrypted remote inference scenario using secure multi-party computation to prevent the disclosure of the data and the model.   

\section{Uncertainty Estimation in FL}
\label{sec:uncertain}

\begin{figure*}[ht]
\centering
\includegraphics[width=\textwidth]{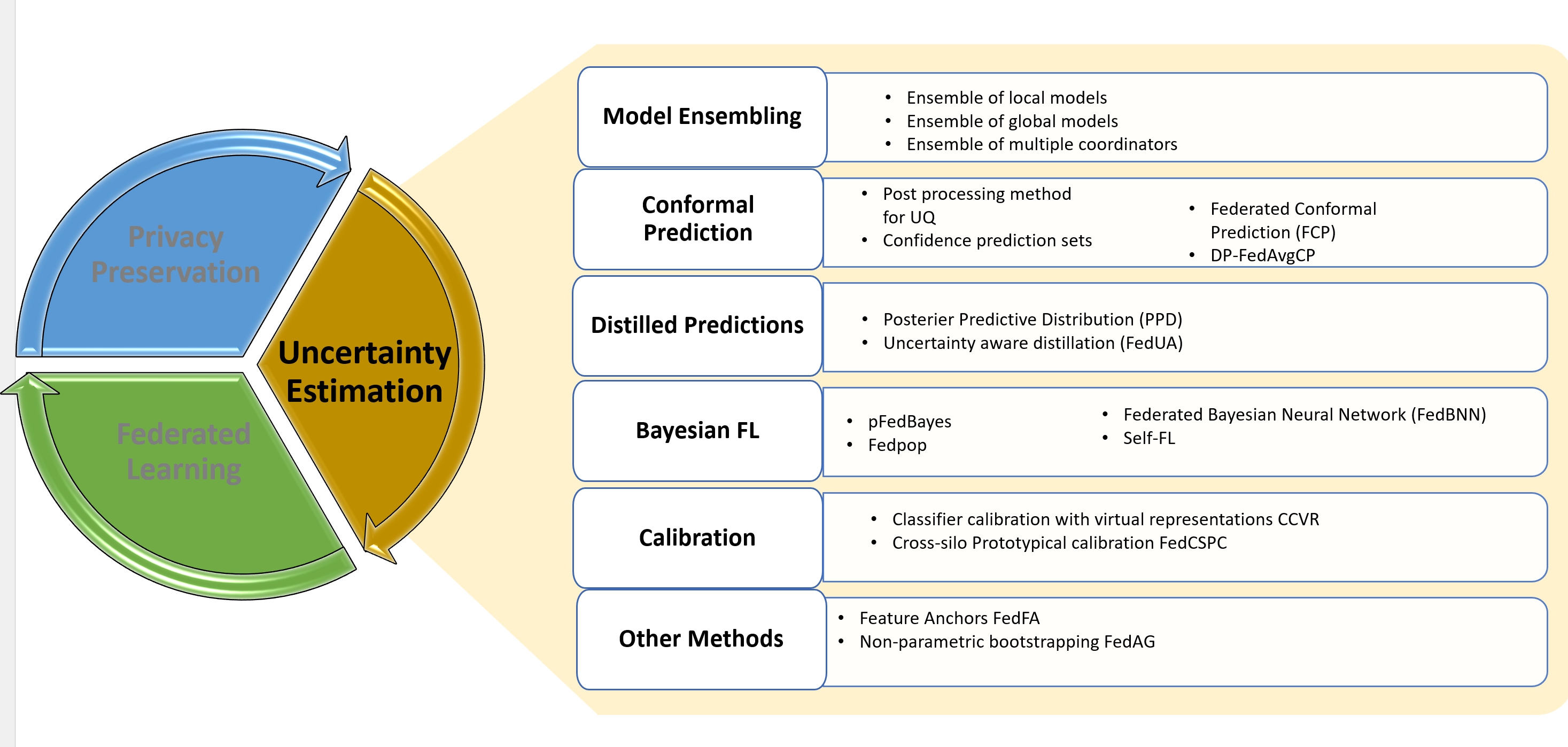} 
\caption{Summary of uncertainty estimation methods in FL.}
\label{fig:5}
\end{figure*}

\begin{table*}[h!]
\centering
\caption{ Uncertainty Estimation Methods in FL.}
\footnotesize{
\begin{tabular}{p{1cm} l p{1cm} p{1cm} p{1cm} p{1cm} p{7cm}}
\hline \hline
\textbf{Algorithm} & \textbf{Ref}  & \textbf{Conformal Prediction (CP)} & \textbf{Distilled Prediction}& \textbf{Bayesian} & \textbf{Calibration} & \textbf{Summary} \\
\hline \hline
Fed-ensemble & \citep{Shi2023fedensemble}  &no  & no & no & no & Extends ensembling methods to FL; characterizes uncertainty in predictions by using the variance in the predictions as a measure of knowledge uncertainty.\\
\hline
 DP-fedCP &\citep{plassier_conformal}   &yes & no & no & no &  Differentially Private Federated Average Quantile Estimation (DP-fedCP); the method is designed to construct personalized CP sets in an FL scenario.\\
\hline
FCP & \citep{Lu2023conformal}  & yes  & no & no & no & Federated CP, a framework for extending CP to FL that addresses the non-i.i.d. nature of data in FL.  \\
\hline
FedPPD & \citep{bhatt2023federated} & no  & yes & no & no & Framework for FL with uncertainty, where, in every round, each client infers the posterior distribution over its parameters
and the posterior predictive distribution (PPD); PPD is sent to the server.\\
\hline
FedUA & \citep{FedUA2023} &no  & yes & no & no & Fed uncertainty aware - Each client’s uncertainty is quantified; a sample quality evaluator selects high-quality samples for global model training; knowledge distillation s used in the aggregation process to transfer inter-class relationships from the local models and suppress noise from incomplete client data.\\
\hline
FedBNN &\citep{makhija2023privacy} &no  & no & yes & no & FL framework based on training a customized local Bayesian model for each client.\\
\hline
pFL & \citep{zhang2023uncertainty} &no  & no & yes & no & The personalized FL (pFL) trains personalized local models to cater to the datasets while still being able to learn from a larger data pool. \\
\hline
Self-FL & \citep{Chen2022SelfAwarePF} &no  & no & yes & no & Self-aware personalized FL method that uses intra-client and inter-client uncertainty estimation to balance the training of its local personal model and global model.\\
\hline
pFedBays &\citep{zhang2022vari} &no  & no & yes & no & Weight uncertainty is introduced in client and server neural networks; to achieve personalization, each client updates its local distribution parameters by balancing its construction error over private data.\\
\hline
Fedpop &\citep{kotelevskii2022pop} &no  & no & yes & no & Each client has a local model composed of fixed population parameters that are shared across clients, as well as random effects that explain heterogeneity in the local data.\\
\hline
FedFA &\citep{zhou2023FA}  &no  & no & no & no & Feature anchors are used to align features and calibrate classifiers across clients simultaneously; this enables client models to be updated in a shared feature space with consistent classifiers during local training. \\
\hline
FedAG & \footnotesize{\citep{Thorgeirsson2020ProbabilisticPW}}) &no  & no & no & no & By introducing weight uncertainty in the aggregation step of FedAvg algorithm, the end devices can calculate probabilistic predictions but only have to learn conventional, deterministic models. \\
\hline
CCVR & \citep{Luo2021NoFO}&no  & no & no & yes & Classifier calibration with Virtual Representation (CCVR) Found a greater bias in representations learned in the deeper layers of a model trained with FL; they show that the classifier contains the greatest bias toward local client data and that classification performance can be greatly improved with post-training classifier calibration calibration\\

\hline
FedCSPC& \citep{QI23silo} &no  & no & no & yes & This method takes additional prototype information from the clients to learn a unified feature space on the server side while maintaining
clear boundaries.\\
\hline \hline
\end{tabular}
}
\label{table:Uncertainty-estimation}
\end{table*}

Another critical area in FL for medical imaging is uncertainty quantification or estimation. Once data privacy is ensured, assessing the quality of the model becomes a crucial focus for researchers. For FL to excel in the medical imaging field, it is essential to have a method for measuring how certain the model is about its predictions. Additionally, there should be a mechanism to alert a human operator when the model's certainty falls below acceptable levels. The unique challenge in FL for medical imaging arises from its non-i.i.d nature, which complicates the quantification of certainty. This complexity is further exacerbated by the phenomenon where local certainty might be high, but global certainty is low, and vice versa. This section will discuss various methods to implement uncertainty estimation in FL settings. Figure \ref{fig:5} and Table \ref{table:Uncertainty-estimation} summarize the uncertainty estimation methods presented in this section.

\subsection{Model Ensembling}
Model ensembling is a popular uncertainty estimation method and involves running inference with an ensemble of models and taking the average \citep{sanderson2018uncertainty}. This naturally extends to FL because of the distributed nature of the FL setup involving multiple clients that can serve as multiple models. The approach in \citep{Linsner2021ApproachesTU} Integrates multiple ensembling methods into an uncertainty estimation framework for FL. The variations of FL ensembling used include \citep{Linsner2021ApproachesTU}:
\begin{itemize}
    \item \textbf{Ensemble of local models}:
    This method is a naive way of incorporating deep ensemble-based uncertainty estimation into FL \citep{Linsner2021ApproachesTU}. This method treats each worker's local model as an ensemble member. Not all the workers communicate with the coordinator, which leads to a $m$ number of separately trained models. These models are then used for final prediction. However, the main idea of FL is lost here due to the lack of communication \citep{Linsner2021ApproachesTU}.
    \item \textbf{Ensemble of global models}: In this approach, the idea of FL is preserved, however computational overhead is increased \citep{Linsner2021ApproachesTU}. Each worker trains $S$ ML models with different random initialization seeds to train each model. For each $S$ model, an FL workflow is executed. This can quickly become computationally expensive as $S$ increases \citep{Linsner2021ApproachesTU}.
    \item \textbf{Ensemble based on multiple coordinators}: These methods split the workers into subgroups and assign a coordinator to each subgroup \citep{Linsner2021ApproachesTU}. FL is carried out as normal among the subgroups, and the outputs of each subgroup are averaged to produce the final prediction.
\end{itemize}

Each method presents advantages and challenges, necessitating careful consideration when used in FL applications in real-world settings.

The ensemble of local models emphasizes privacy and simplicity by treating each worker's model as an independent ensemble member. While this approach maximizes data privacy and is straightforward to implement, it diverges from the collaborative essence of FL. It may result in inconsistent model performance due to isolated training environments. Conversely, the ensemble of global models aligns with the collaborative learning principle of FL, enhancing model robustness by integrating diverse perspectives. However, this method significantly increases computational and communication demands, posing scalability challenges as the number of clients grows.
The third approach, employing multiple coordinators, offers improved scalability by distributing the workload and tailoring learning strategies within subgroups. However, this method introduces additional complexity in coordination and risks learning fragmentation across subgroups.
To navigate these trade-offs, considering hybrid or adaptive ensembling strategies that balance computational efficiency with the benefits of collaborative learning could be beneficial. Ultimately, selecting an ensembling method should be guided by the application's specific needs, including privacy requirements, available computational resources, and data heterogeneity.

\subsubsection{Fed-ensemble}
Fed-ensemble \citep{Shi2023fedensemble} extends ensembling methods for FL using a different approach. Instead of aggregating local models to update a single global model, this method uses random permutations to update a group of $K$ models and obtains predictions using model averaging. This method imposes no additional computational costs and can readily be utilized within established FL algorithms. The authors empirically show that the proposed approach performs superior to other methods on many datasets. It also excels in heterogeneous settings, which is consistent with many FL applications like medical imaging. Fed-ensemble can characterize uncertainty in predictions by using the variance in the predictions as a measure of knowledge uncertainty. Shi et. al \citep{Shi2023fedensemble} propose performing ensemble FL that updates $K$ models over local datasets. Point predictions are obtained by model averaging. The authors show that the Fed-ensemble excels at uncertainty quantification when tested on CIFAR-10 \citep{cifar10} CIFAR-100 \citep{CIFAR}, MNIST, and the Openimagesv4 dataset \citep{Kuznetsova2020OpenImagesV4} in both homogeneous and heterogeneous settings. Using  NTK, they show that predictions at new data points from all $K$ models converge to samples from the same, limiting the Gaussian process in sufficiently over-parameterized regimes \citep{Shi2023fedensemble}. The server sends one of the $K$ models to every client in each training round to train on local data. The server then aggregates the updated model from each client; this way, the burden on clients is not increased, and all $K$ models eventually see all the clients' data. To obtain uncertainty predictions in an ensemble of models, the sample variance can be used to estimate the uncertainty. Fed-ensemble can appropriately characterize knowledge uncertainty on regions without labeled data. Fed-ensemble enhances existing FL techniques by systematically quantifying uncertainty and increasing model capacity without raising communication costs. Unlike Fedavg, which tends to be overconfident in predictions, Fed-ensemble offers convergence guarantees and effectively manages data heterogeneity through ensembling, outperforming methods that rely on strong regularizers.

\subsection{Conformal Prediction (CP)}
CP is another method for uncertainty estimation that has been extensively explored in FL. The idea was first proposed in \citep{gammerman98conf} and then improved upon by \citep{saunders99conf} around the turn of the century. CP is a statistical framework that is used to provide reliable and valid confidence measures for the predictions made by ML models. CP begins by defining a nonconformity measure, which quantifies how different a new example is from a set of previously seen examples \citep{saunders99conf}. This measure is based on an ML algorithm trained on a dataset. The non-conformity of an example can be something like the distance from a decision boundary in classification or the error of a prediction in regression. 

For a new data sample, CP generates prediction regions (or sets) likely to contain the true label or value. This is done by considering all possible labels for the new example, calculating the nonconformity score for each label, and comparing these scores to the scores from the calibration set. Lu et al. \citep{Lu2021DistributionFreeFL} point out that since CP is primarily a post-processing method for uncertainty estimation, integrating it into an FL framework is generally straightforward. The authors also correlate class entropy with prediction set size to determine task uncertainty. CP can produce confidence predictions for any ML model that outputs a score function. 

\subsubsection{Differentialy Private Federated Average Quantile Estimation (DP-FedAvgQE)} 
DP-FedAvgQE brings CP and DP to FL and provides theoretical privacy guarantees to ensure additional security  \citep{plassier_conformal}. DP-FedAvgQE provided strong benchmarks on ImageNet  \citep{ImageNet} and CIFAR-10 \citep{cifar10} datasets and simulated data experiments. DP-FedAvgQE takes advantage of importance weighting to address the label shift between agents effectively. 

\subsubsection{Federated Conformal Prediction (FCP)} 
FCP is another method for extending CP to FL that addresses the non-i.i.d. nature of data in FL\citep{Lu2023conformal}.  The inherent heterogeneity of FL datasets violates the fundamental tenet of exchangeability between the calibration data distribution and the test data distribution during inference in CP, implying that the calibration and test data have identical distributions \citep{vovk2005algorithmic}. To solve this violation, the authors propose using partial exchangeability, which is a generalization of exchangeability \citep{Lu2023conformal}. FCP makes no assumptions between clients  $P1, . . . , PK.$.  Specifically,
this assumption does not require independence or identical distributions among clients. FCP provides rigorous theoretical guarantees and excellent empirical performance on multiple computer vision and medical imaging datasets.

\subsection{Distilled Predictions}
The distilled prediction method leverages knowledge distillation to quantify uncertainty \citep{bhatt2023federated}.

\subsubsection{Federated Posterior Predictive Distribution (FedPPD)} 
FedPPD is a framework for FL with uncertainty estimation where, in every round, each client infers the posterior distribution over its parameters and the posterior predictive distribution (PPD) \citep{bhatt2023federated}. The estimated PPD is sent to the server. Making predictions at test time does not require computationally expensive Monte-Carlo averaging over the posterior distribution because this approach maintains the PPD in the form of a single deep neural network. Moreover, this approach makes no restrictive assumptions, such as the form of the clients’ posterior distributions or their PPDs. FedPPD follows a two-step process \citep{bhatt2023federated}:
\begin{itemize}
    \item \textbf{Step 1} -  For each client, the authors perform approximate Bayesian inference for the posterior distribution of the client model weights using Markov Chain Monte Carlo (MCMC) sampling \citep{bhatt2023federated}. This produces a set of samples from the client’s posterior, and these samples are used as teacher models, which are distilled into a student model. The authors use stochastic gradient Langevin dynamics (SGLD) sampling since it provides an online method to distill these posterior samples efficiently into a student model.
    \item \textbf{Step 2} - For each client, the authors distill the MCMC samples (teacher models) directly into the PPD, which is the student model \citep{bhatt2023federated}. Notably, in this distillation-based approach, the PPD for each client is represented succinctly by a single deep neural network instead of via an ensemble of deep neural networks. This makes the prediction stage much faster than typical Bayesian approaches.
\end{itemize}

\subsubsection{Fed Uncertainty Aware FedUA} 
FedUA provides another approach to distill predictions focusing on non-i.i.d. data while limiting communication bandwidth \citep{FedUA2023}. This framework implements two core components: (1) uncertainty measurement to quantify each client's uncertainty and (2) a sample quality evaluator to select high-quality samples for global model training. Knowledge distillation is used in the aggregation process to transfer inter-class relationships from the local models and suppress noise from incomplete client data. The authors empirically show that FedUA improves accuracy compared to other FL models while limiting communication costs on image classification tasks. The authors also reported that the uncertainty measurement using feature space density was more robust to native data uncertainty than softmax entropy. 

Knowledge distillation provides two promising improvements for FL: it can alleviate overfitting on client-side data by sifting through informative and valuable information for learning, mitigating the bias caused by incomplete or over-trained data on a given client. Knowledge distillation also allows the global model to learn inter-class relationships, which helps to transfer knowledge from a general multi-purposed model to a specific target-oriented model. For the sample evaluator, the method used when finding quality samples can be described as ``samples that do not reach consensus among local models should be taken with a higher priority''. These samples are more important for optimizing the local model on the server side. For uncertainty estimation,  a single deterministic model is used to quantify uncertainty by estimating feature space density for each client model. For a new input sample, features are extracted and evaluated to get the probability density function in the client's feature space. A lower density indicates a higher uncertainty and vice versa. Leveraging knowledge distillation is a powerful way of implementing uncertainty estimation into an FL framework with non-i.i.d data \citep{FedUA2023}.

\subsection{Bayesian FL}
A popular class of models for providing uncertainty quantification in ML belongs to Bayesian or probabilistic models \citep{brown2022simple}. These models utilize Bayesian methods to give probabilistic predictions rather than point predictions \citep{swaminathan2022bayesian}. 

\subsubsection{Federated Bayesian Neural Network (FedBNN)}
The probabilistic predictions can give insight into model uncertainty \citep{makhija2023privacy}. The authors present a unified FL framework based on training a customized local Bayesian model for each client. These models can learn in the absence of large local datasets. The Bayesian nature of these models allows for incorporating supervision in the form of prior distributions. The authors use the prior of the functional output space of the network to aid in collaboration across heterogeneous clients \citep{makhija2023privacy}. 

\subsubsection{Personalized FL (pFL)}
In some settings with heterogeneous data, it makes sense to personalize local models to cater to their respective datasets while still being able to learn from a larger data pool, like the work done by \citep{DALMAZ2024103121} for multi-contrast MRI synthesis. This practice is known as personalized FL (pFL) and can be carried out in two primary ways with Bayesian techniques according to \citep{zhang2023uncertainty}:
\begin{itemize}
     \item \textbf{Global model personalization}: The global model personalization strategy begins with the training of a global model on data distributed across many devices or nodes. The model is trained by aggregating locally computed updates from each node without sharing the data itself. Once this global model has been trained, it can be personalized for individual users or clients.
    \item \textbf{Personalized model learning}: With personalized model learning in a federated setting, the focus shifts to training individual models for each site from the outset, leveraging the local data while still occasionally sharing insights or parameters (in a privacy-preserving manner) across the network to improve the models collectively.
 \end{itemize}
 
The study by Zhang et al. \citep{zhang2023uncertainty} shows that personalization in FL improves classification accuracy and increases the quality of estimated uncertainty \citep{zhang2023uncertainty}. Thus, personalization is a promising research direction in local client deployment and uncertainty quantification for healthcare applications \citep{zhang2023uncertainty}. Bayesian methods are heavily used for creating pFL algorithms \citep{zhang2023uncertainty}. 

\subsubsection{Self-FL} 
Self-aware personalized FL (Self-FL) uses intra-client and inter-client uncertainty to balance the training of local personal and global models \citep{Chen2022SelfAwarePF}. Larger inter-client variation implies more personalization is needed. Self-FL uses uncertainty-driven local training steps and aggregation rules instead of conventional local fine-tuning and sample size-based aggregation. The authors interpret personalized FL through a two-level Bayesian hierarchical model perspective to characterize client-specific and globally-shared information. The method uses uncertainty to drive client-side training with an adaptive number of local steps and server-side aggregation (variance-weighted averaging). The authors evaluate their method using synthetic data, images (MNIST, FEMNIST \citep{femnist}, CIFAR10 \citep{cifar10}), text (Sent140 \citep{sent140}), and audio (wake-word detection) and show robust personalization capability under data heterogeneity. Some key advantages of the Self-FL model are principled connections to hierarchical Bayesian modeling and built-in auto-tuning of hyper-parameters for each client, all while maintaining the same computation and communication overhead as FedAvg \citep{Chen2022SelfAwarePF}. 

\subsubsection{Personalized federated learning with Bayesian inference (pFedBays)}
In pFedBays, weight uncertainty is introduced in neural networks for clients and the server \citep{zhang2022vari}. To achieve personalization, each client updates its local distribution parameters by balancing its construction error over private data and its Kullback–Leibler (KL) divergence with global probability distribution from the server. pFedBays method tackles two issues in FL: training on non-i.i.d data across clients and overfitting due to limited data \citep{zhang2022vari}. pFedBays formulates both the local clients' models and the global server model as Bayesian neural networks, where the parameters are modeled as probability distributions rather than point estimates. This helps address overfitting.
The server optimizes to find a global distribution that is close to the local distributions by minimizing KL divergence. Each client balances minimizing a local data fit term and the KL divergence from the global distribution to find its personalized distribution. The global distribution acts as a prior for the local models. By replacing the prior distribution with a trained global distribution, the authors find a relatively good distribution without making assumptions about the prior distribution \citep{zhang2022vari}. This is critical because estimating a prior in many real-world scenarios is not feasible \citep{zhang2022vari}. The authors provide a theoretical analysis bounding the generalization error and showing the convergence rate is minimax optimal up to a logarithmic factor. Being a Bayesian network, uncertainty estimates for the parameters can be monitored to understand the model's confidence in its predictions. 

\subsubsection{Fedpop}
The Fedpop \citep{kotelevskii2022pop} framework recasts the method of pFL into a population modeling paradigm. Clients integrate fixed common population parameters with random effects, expanding data heterogeneity. Each client has a local model composed of fixed population parameters shared across clients and random effects that explain heterogeneity in the local data. Kotelevskii et al. developed a new stochastic optimization algorithm based on MCMC to perform inference under this model \citep{kotelevskii2022pop}. The algorithm allows uncertainty estimation, handles issues like client drift, and works well even with limited client participation. The authors show that, in practice, the added computational cost from the Monte Carlo chain is negligible. FedPop allows for uncertainty estimation by having each client model involve a fixed effect parameter shared across clients and a low-dimensional random effect parameter sampled for each client. Introducing a common prior on the local parameters addresses the local overfitting problem where clients have highly heterogeneous and small datasets\citep{kotelevskii2022pop}.

\subsection{Other methods for Uncertainty Estimation in FL}
Some methods take a different approach to quantifying uncertainty in FL setups.

\subsubsection{Feature Anchors}
Zhou et al. propose using feature anchors to align features and classifiers for heterogeneous data in their framework \citep{zhou2023FA}. FedFA is designed to address the challenges posed by heterogeneous data. This method utilizes feature anchors to align features and calibrate classifiers across clients simultaneously. This enables client models to be updated in a shared feature space with consistent classifiers during local training. The authors explain the vicious and virtuous cycles of FL with heterogeneous data:
\begin{itemize}
    \item \textbf{Vicious Cycle}: In traditional FL approaches, data heterogeneity leads to feature inconsistency across client models. This inconsistency causes classifier updates to diverge, forcing feature extractors to map to more inconsistent feature spaces. This cycle of increasing divergence in classifiers and inconsistency in features degrades the performance and convergence of the federated model.
    \item \textbf{Virtuous Cycle}: FedFA introduces feature anchors to break this vicious cycle. By aligning features and calibrating classifiers across clients, FedFA creates a virtuous cycle. The aligned features and classifiers promote consistency in client features and classifiers. This alignment ensures that client models are updated in a shared, consistent feature space with similar classifiers, leading to improved performance and more stable convergence.
\end{itemize}

The FedFA framework integrates feature anchor loss to minimize local objective functions. This mechanism is designed to align class-specific features and diminish the distance within classes at the client level \citep{zhou2023FA}. Moreover, the FedFA algorithm encompasses a server-side component where both class feature anchors and the global model undergo aggregation. This process employs a weighted averaging scheme akin to that of the FedAvg algorithm, facilitating the integration of local updates into a cohesive global model.

\subsubsection{FedAvg-Gaussian FedAG}
FedAvg-Gaussian FedAG takes a Gaussian approach to generating probabilistic predictions in FL \citep{Thorgeirsson2020ProbabilisticPW}. By introducing weight uncertainty in the aggregation step of the FedAvg algorithm, the end devices can calculate probabilistic predictions but only have to learn conventional, deterministic models. This allows for uncertainty estimation in an FL framework. The key idea in FedAG is to treat the probability distribution of the local model weights from different devices as an approximation of the posterior distribution over the global model weights \citep{Thorgeirsson2020ProbabilisticPW}. This allows the global model to make probabilistic predictions. For linear models, FedAG performs on par with Bayesian linear regression. For neural networks, FedAG outperforms variational inference methods and approaches the performance of deep ensembles for probabilistic predictions after several rounds of training \citep{Thorgeirsson2020ProbabilisticPW}. FedAG provides an efficient and privacy-preserving way to enable probabilistic predictions in FL settings, with performance competitive to non-federated methods. FedAG has comparable accuracy to non-FL and non-distributed learning frameworks. There are two variations to this algorithm: 
\begin{itemize}
    \item \textbf{Monte Carlo and non-parametric bootstrapping}: In Monte Carlo and non-parametric bootstrapping, $M$ sets of weights are randomly drawn from the posterior weight distributions learned during federated aggregation. Each of these $M$ weight sets is used to generate a prediction on the test input. These $M$ predictions are aggregated (by taking mean and variance) to approximate a predictive distribution. 
    \item \textbf{Non-parametric bootstrapping}: This approach uses the local weight updates from client devices directly, rather than drawing samples from the fitted posterior probability distributions, to generate predictions. Each client's weight update is used directly to generate a prediction. These predictions approximate the predictive distribution. Non-parametric bootstrapping is conceptually similar to bootstrap aggregating \citep{breiman1996bagging}, where re-sampling the training data is replaced by re-sampling the client weights. 
\end{itemize}
    
FedFA used its feature anchors to calibrate the model's classifier. Calibration is a different way of dealing with model uncertainty, and it will be discussed in the next part of paper.

\subsection{Calibration}
Calibration is another method of dealing with uncertainty estimation by correcting an ML model's tendency to be overconfident in incorrect predictions due to the Softmax function \citep{pearce2021understanding}. By calibrating the confidence, a better assumption about the quality of a prediction can be made \citep{pearce2021understanding}. 

\subsubsection{Classifier Calibration with Virtual Representation (CCVR)}
CCVR calibrates a global model to improve performance on non-i.i.d data in heterogeneous settings \citep{Luo2021NoFO}. The authors found a greater bias in representations learned in the deeper layers of a model trained with FL. They show that the classifier contains the greatest bias and that post-calibration can greatly improve classification performance. Specifically, the classifiers learned on different clients show the lowest feature similarity. The classifiers tend to get biased toward the classes over-represented in the local client data, leading to poor performance in under-represented classes. This classifier bias is a key reason behind performance degradation on non-i.i.d federated data. Regularizing the classifier during federated training brings minor improvements \citep{Luo2021NoFO}. However, post-training calibration of the classifier significantly improves classification accuracy across various FL algorithms and datasets \citep{Luo2021NoFO}. CCVR generates virtual representations using Gaussian probability distributions fitted on client feature statistics. CCVR then retrains the classifier on these virtual representations while fixing the feature extractor. Experimental results show state-of-the-art accuracies on common benchmark datasets like CIFAR-10. CCVR is built on top of the off-the-shelf feature extractor and requires no transmission of the representations of the original data, thus raising no additional privacy concerns. 

\subsubsection{FedCSPC} 
FedCSPC method addresses the issue of heterogeneous data distributions across clients in FL \citep{QI23silo}. This method takes additional prototype information from the clients to learn a unified feature space on the server side while maintaining clear boundaries. There are two main modules to this framework: (1) The Data Prototypical Modeling (DPM) module and (2) the Cross-Silo Prototypical Calibration (CSPC) module \citep{QI23silo}. The DPM module uses clustering to model representation distributions for each client and generate class-specific prototypes for the server. This helps capture diversity within each class. The CSPC module on the server aligns the heterogeneous prototype features from different clients into a unified space. It uses an augmented contrastive learning approach with positive mixing and hard negative mining to improve robustness. Knowledge-based predictions using the calibrated exemplar prototypes from the unified space to supplement the network predictions. FedCSPC alleviates the feature gap between data sources, which can significantly improve generalization ability.The authors test the proposed framework on CIFAR10, CIFAR100 \citep{CIFAR}, TinyImageNet \citep{tinyimagenet}, and VireoFood172 \citep{vireofood172} datasets. The proposed CSPC module is an orthogonal improvement to client-based methods and has a plug-and-play design that makes it easy to integrate into existing FL frameworks. Calibration is an attractive method for FL as it introduces little additional communication overhead and can effectively provide quality information about model certainty \citep{QI23silo}. 

\section{Real-World Applications}
\label{sec:applications}

With the influx of research in the field of FL for medical imaging, some successful real-world applications showcase FL's potential for the medical imaging domain \citep{pati2021federated}. The Federated Tumor Segmentation (FeTS-1.0) Challenge \citep{pati2021federated} was the first real-world FL challenge for medical images. The goals for this challenge were:
\begin{itemize}
    \item The identification of the optimal weight aggregation approach towards training a consensus model that has gained knowledge via FL from multiple geographically distinct institutions while their data are always retained within each institution.
    \item The federated evaluation of the generalizability of brain tumor segmentation models ``in the wild'', i.e., on data from institutional distributions that were not part of the training datasets.
\end{itemize}
The FeTS-1.0 study opened the door for FL in a medical environment. Participants were given a U-net \citep{ronneberger2015unet} model and tasked with finding the best method for weight aggregation. The FeTS-1.0 challenge also focused on the real-world evaluation of the consensus model to show if it could perform well on real unseen data. The success of this first challenge paved the way for the FeTS-2.0 Challenge \citep{pati2022federated}, where the objective was to address out-of-sample generalizability for rare Glioblastoma cancer boundary detection. Due to this disease's rarity and privacy concerns regarding medical data, it is a challenge to gather a large amount of data to train a model on this task \citep{pati2022federated}. Traditional approaches to this problem involve sharing multi-site data \citep{pati2022federated}, but centralizing such data is often difficult or infeasible due to various limitations regarding privacy. The study presented in this paper is the largest FL project to date, incorporating data from $71$ sites around the globe  \citep{pati2022federated}. With this approach, the authors created the largest Glioblastoma dataset with $6,214$ samples. 
The authors reported a $33\%$  improvement in segmenting the surgically targetable portion of the tumor and a $23\%$  improvement for the complete tumor compared to a publicly trained model  \citep{pati2022federated}. 
This research demonstrates that FL enhances the efficacy of ML methodologies within the medical sector, reinforcing the notion that FL could be a transformative technology for amplifying the impact of ML in medical imaging.

\section{Challenges and Opportunities}
\label{sec:challenges}

While there has been significant progress in FL in recent years, some challenges remain to be solved. These challenges present potential opportunities for researchers to further explore and improve the state of FL for medical imaging. One particular challenge is the inherent trade-off between privacy and security in FL. Further research into the efficient allocation of the privacy budget to enhance model performance without compromising privacy is a key area that requires further research. In addition to privacy and security, communication efficiency must also be considered. Alternate noise addition methods are also a possible route for increasing the effectiveness of DP, as current methods may not be optimal. Another trade-off that still presents a challenge is personalization versus overfitting. Personalization in FL can increase accuracy but risks affecting uncertainty estimation performance due to overfitting. Methods for optimizing personalization to balance overfitting are open areas for research. Computational efficiency remains an issue for many aspects of FL, particularly with deep ensembles like fed-ensemble. Finding more computationally efficient methods could progress FL further. uncertainty estimation for out-of-distribution and noisy labels is an under-researched area, and there is a need to investigate how uncertainty estimation can be leveraged to address these issues. Exploring generative AI models to provide application-specific alignment datasets could be a promising direction. Generative AI could make up for a lack of data by providing simulated data. Conformal prediction has been shown to perform well for uncertainty estimation in FL, but little research has been conducted on conformal prediction in a personalized setting, making it an open research area. Ensemble modes have been integrated into FL and could potentially address and detect client drift, anomalies, or fairness challenges during model training. Applying data-free knowledge transfer methods could improve practicability in scenarios where shared datasets are not available, providing a secure way to transfer knowledge across clients.

\section{Conclusion}
\label{sec:conclusion}

Machine learning holds the potential to dramatically improve the effectiveness of medical imaging for disease diagnoses and treatment. However, to succeed, methods need to be implemented to address both privacy concerns and uncertainty estimation. FL is a powerful solution for training on multiple private datasets without exposing any private data, and enhanced privacy preservation and uncertainty estimation methods can be an effective approach for training large medical imaging models. This paper provided a comprehensive review of the current state of FL algorithms, privacy preservation, and uncertainty estimation in the context of medical imaging. Significant progress has been made in recent years to make FL viable for the medical imaging domain, with work being done to optimize the aggregation process, privacy preservation, and uncertainty estimation.

\section{Acknowledgements}
This work was funded by NSF grants 2234468 and 2234836. The content is the responsibility of the authors and does not reflect the official views of the National Science Foundation
%%Harvard
\bibliography{references.bib} 

\begin{thebibliography}{100}

\bibitem{Erickson2017}
B.~Erickson, P.~Korfiatis, Z.~Akkus, and T.~Kline, ``Machine learning for medical imaging,'' {\em Radiographics}, 2017.

\bibitem{Latif2019}
J.~Latif, C.~Xiao, A.~Imran, and S.~Tu, ``Medical imaging using machine learning and deep learning algorithms: A review,'' in {\em ICOMET}, 2019.

\bibitem{BarraganMontero2021}
A.~M. Barragán-Montero {\em et~al.}, ``Artificial intelligence and machine learning for medical imaging: A technology review,'' {\em Physica Medica}, 2021.

\bibitem{Willemink2020}
M.~Willemink {\em et~al.}, ``Preparing medical imaging data for machine learning,'' {\em Radiology}, 2020.

\bibitem{pmid24892406}
J.~Jager, T.~Gremeaux, T.~Gonzalez, S.~Bonnafous, C.~Debard, M.~Laville, and H.~Vidal, ``Adipose tissue-derived stem cells promote monocyte recruitment in adipose tissue and liver,'' {\em Mol Metab}, vol.~3, no.~4, pp.~417--425, 2014.

\bibitem{pmid26579733}
C.~Wu, H.~Ying, F.~Grinnell, G.~Bryant-Greenwood, R.~Riha, J.~Nguyen, Z.~Li, M.~Parsons, B.~Parry, D.~Rotstein, A.~Lightfoot, and S.~Cassar, ``Vitamin d receptor localization and activity in early human fetal development,'' {\em J Clin Endocrinol Metab}, vol.~100, no.~12, pp.~E1568--E1575, 2015.

\bibitem{pmid30720861}
G.~Sahay, D.~Y. Alakhova, and A.~V. Kabanov, ``Endocytosis of nanomedicines,'' {\em J Control Release}, vol.~145, no.~3, pp.~182--195, 2010.

\bibitem{pmid32917666}
N.~Bettencourt, J.~P. Ferreira, I.~P. Culotta, J.~J. McMurray, S.~Jacob, J.~L. Rouleau, K.~Swedberg, S.~J. Pocock, S.~D. Solomon, F.~Zannad, and P.~Rossignol, ``Impact of intensive versus standard blood pressure lowering in chronic kidney disease patients with and without diabetes mellitus: A subanalysis of the sprint study,'' {\em Hypertension}, vol.~76, no.~4, pp.~979--987, 2020.

\bibitem{DAYARATHNA2024103046}
S.~Dayarathna, K.~T. Islam, S.~Uribe, G.~Yang, M.~Hayat, and Z.~Chen, ``Deep learning based synthesis of mri, ct and pet: Review and analysis,'' {\em Medical Image Analysis}, vol.~92, p.~103046, 2024.

\bibitem{HIPAA1996}
``{Health Insurance Portability and Accountability Act of 1996}.'' Pub. L. No. 104-191, 110 Stat. 1936, 1996.
\newblock Available from: U.S. Government Printing Office, via: \url{https://www.govinfo.gov/content/pkg/PLAW-104publ191/pdf/PLAW-104publ191.pdf}.

\bibitem{GDPR2016}
``{General Data Protection Regulation (GDPR)}.'' Regulation (EU) 2016/679 of the European Parliament and of the Council of 27 April 2016, 2016.
\newblock Available from: \url{https://eur-lex.europa.eu/eli/reg/2016/679/oj}.

\bibitem{pmlr-v54-mcmahan17a}
B.~McMahan, E.~Moore, D.~Ramage, S.~Hampson, and B.~A.~y. Arcas, ``{Communication-Efficient Learning of Deep Networks from Decentralized Data},'' in {\em Proceedings of the 20th International Conference on Artificial Intelligence and Statistics} (A.~Singh and J.~Zhu, eds.), vol.~54 of {\em Proceedings of Machine Learning Research}, pp.~1273--1282, PMLR, 20--22 Apr 2017.

\bibitem{Qu2021heter}
L.~Qu, N.~Balachandar, and D.~Rubin, ``An experimental study of data heterogeneity in federated learning methods for medical imaging,'' {\em arXiv}, 2021.
\newblock Available at arXiv:2107.08371.

\bibitem{Max2021Gdisag}
M.~Lam, G.~Wei, D.~Brooks, V.~J. Reddi, and M.~Mitzenmacher, ``Gradient disaggregation: Breaking privacy in federated learning by reconstructing the user participant matrix,'' {\em CoRR}, vol.~abs/2106.06089, 2021.

\bibitem{Wu2022invert}
R.~Wu, X.~Chen, C.~Guo, and K.~Q. Weinberger, ``Learning to invert: Simple adaptive attacks for gradient inversion in federated learning,'' {\em arXiv}, 2022.

\bibitem{Jere2021taxonomy}
M.~Jere, T.~Farnan, and F.~Koushanfar, ``A taxonomy of attacks on federated learning,'' {\em IEEE Security \& Privacy}, 2021.

\bibitem{Dwork2006}
C.~Dwork, ``Differential privacy,'' in {\em 33rd International Colloquium on Automata, Languages and Programming, part II (ICALP 2006)}, pp.~1--12, Springer, 2006.

\bibitem{Gentry2009}
C.~Gentry, {\em A Fully Homomorphic Encryption Scheme}.
\newblock PhD thesis, Stanford University, 2009.

\bibitem{Linsner2021ApproachesTU}
F.~Linsner, L.~Adilova, S.~D{\"a}ubener, M.~Kamp, and A.~Fischer, ``Approaches to uncertainty quantification in federated deep learning,'' in {\em PKDD/ECML Workshops}, 2021.

\bibitem{Psaros2022}
A.~F. Psaros, X.~Meng, Z.~Zou, L.~Guo, and G.~Karniadakis, ``Uncertainty quantification in scientific machine learning: Methods, metrics, and comparisons,'' {\em Journal of Computational Physics}, 2022.

\bibitem{Darzidehkalani2022}
E.~Darzidehkalani, M.~Ghasemi-rad, and P.~V. van Ooijen, ``Federated learning in medical imaging: Part ii: Methods, challenges, and considerations,'' {\em Journal of the American College of Radiology}, vol.~19, no.~4, pp.~P755--765, 2022.

\bibitem{kaissis2020secure}
G.~Kaissis, M.~Makowski, D.~Rückert, and R.~Braren, ``Secure, privacy-preserving and federated machine learning in medical imaging,'' {\em Nature Machine Intelligence}, vol.~2, 06 2020.

\bibitem{Mouhni2022}
N.~Mouhni, A.~Elkalay, M.~Chakraoui, A.~Abdali, A.~Ammoumou, and I.~Amalou, ``Federated learning for medical imaging: An updated state of the art,'' {\em Ingenierie des Systemes d'Information}, vol.~27, no.~1, pp.~117--122, 2022.

\bibitem{li2020federated}
T.~Li, A.~K. Sahu, M.~Zaheer, M.~Sanjabi, A.~Talwalkar, and V.~Smith, ``Federated optimization in heterogeneous networks,'' 2020.

\bibitem{Li2021FedBNFL}
X.~Li, M.~Jiang, X.~Zhang, M.~Kamp, and Q.~Dou, ``Fedbn: Federated learning on non-iid features via local batch normalization,'' {\em ArXiv}, vol.~abs/2102.07623, 2021.

\bibitem{Yu2022tct}
Y.~Yu, A.~Wei, S.~Karimireddy, Y.~Ma, and M.~Jordan, ``Tct: Convexifying federated learning using bootstrapped neural tangent kernels,'' 07 2022.

\bibitem{lu2022personal}
W.~Lu, J.~Wang, Y.~Chen, X.~Qin, R.~Xu, D.~Dimitriadis, and T.~Qin, ``Personalized federated learning with adaptive batchnorm for healthcare,'' {\em IEEE Transactions on Big Data}, pp.~1--1, 2022.

\bibitem{zhu2021dfree}
Z.~Zhu, J.~Hong, and J.~Zhou, ``Data-free knowledge distillation for heterogeneous federated learning,'' {\em Proceedings of machine learning research}, vol.~139, pp.~12878--12889, 07 2021.

\bibitem{Liu2021ABF}
L.~Liu, X.~Jiang, F.~Zheng, H.~Chen, G.-J. Qi, H.~Huang, and L.~Shao, ``A bayesian federated learning framework with online laplace approximation.,'' {\em IEEE transactions on pattern analysis and machine intelligence}, vol.~PP, 2021.

\bibitem{huang2022learn}
W.~Huang, M.~Ye, and B.~Du, ``Learn from others and be yourself in heterogeneous federated learning,'' in {\em 2022 IEEE/CVF Conference on Computer Vision and Pattern Recognition (CVPR)}, pp.~10133--10143, 2022.

\bibitem{warnat2021swarm}
S.~Warnat-Herresthal, H.~Schultze, K.~Shastry, S.~Manamohan, S.~Mukherjee, V.~Garg, R.~Sarveswara, K.~Händler, P.~Pickkers, N.~A. Aziz, M.~Breteler, E.~Giamarellos-Bourboulis, M.~Kox, M.~Becker, S.~Cheran, M.~Woodacre, E.~Goh, J.~Schultze, and H.~Grundmann, ``Swarm learning for decentralized and confidential clinical machine learning,'' 01 2021.

\bibitem{Kalra2023Proxy}
S.~Kalra, J.~Wen, J.~C. Cresswell, M.~Volkovs, and H.~R. Tizhoosh, ``Decentralized federated learning through proxy model sharing,'' {\em Nature Communications}, vol.~14, p.~2899, May 2023.

\bibitem{butt2023fog}
M.~Butt, N.~Tariq, M.~Ashraf, H.~S. Alsagri, S.~A. Moqurrab, H.~A.~A. Alhakbani, and Y.~A. Alduraywish, ``A fog-based privacy-preserving federated learning system for smart healthcare applications,'' {\em Electronics}, vol.~12, no.~19, 2023.

\bibitem{Wu2021}
J.~Wu, Q.~Xia, and Q.~Li, ``Efficient privacy-preserving federated learning for resource-constrained edge devices,'' in {\em IEEE International Conference on Mobile Ad-Hoc and Smart Systems (MASS)}, 2021.

\bibitem{Mills2020}
J.~Mills, J.~Hu, and G.~Min, ``Communication-efficient federated learning for wireless edge intelligence in iot,'' {\em IEEE Internet of Things Journal}, vol.~7, no.~7, pp.~5986--5994, 2020.

\bibitem{sheller2020fed}
M.~Sheller, B.~Edwards, G.~Reina, J.~Martin, S.~Pati, A.~Kotrotsou, M.~Milchenko, W.~Xu, D.~Marcus, R.~Colen, and S.~Bakas, ``Federated learning in medicine: facilitating multi-institutional collaborations without sharing patient data,'' {\em Scientific Reports}, vol.~10, 07 2020.

\bibitem{wen2022survey}
J.~Wen, Z.~Zhang, Y.~Lan, Z.-s. Cui, J.~Cai, and W.~Zhang, ``A survey on federated learning: challenges and applications,'' {\em International Journal of Machine Learning and Cybernetics}, 2022.

\bibitem{Liu2022TowardsMO}
D.~Liu, L.~Bai, T.~Yu, and A.~Zhang, ``Towards method of horizontal federated learning: A survey,'' {\em 2022 IEEE 5th International Conference on Big Data and Artificial Intelligence (BDAI)}, 2022.

\bibitem{Leroy2019Federated}
D.~Leroy, A.~Coucke, T.~Lavril, T.~Gisselbrecht, and J.~Dureau, ``Federated learning for keyword spotting,'' in {\em Proceedings of IEEE International Conference on Acoustics, Speech and Signal Processing}, pp.~6341--6345, 2019.

\bibitem{Ramaswamy2019Federated}
S.~Ramaswamy, R.~Mathews, K.~Rao, and F.~Beaufays, ``Federated learning for emoji prediction in a mobile keyboard,'' {\em arXiv preprint arXiv:1906.04329}, 2019.

\bibitem{Fallah2020Personalized}
A.~Fallah, A.~Mokhtari, and A.~Ozdaglar, ``Personalized federated learning with theoretical guarantees: A model-agnostic meta-learning approach,'' in {\em Advances in Neural Information Processing Systems}, vol.~33, 2020.

\bibitem{Zhang2022Challenges}
K.~Zhang, X.~Song, C.~Zhang, and C.~Yu, ``Challenges and future directions of secure federated learning: a survey,'' {\em Frontiers in Computer Science}, vol.~16, no.~5, p.~165817, 2022.

\bibitem{Liu2022VerticalFL}
Y.~Liu, Y.~Kang, T.~Zou, Y.~Pu, Y.~He, X.~Ye, Y.~Ouyang, Y.~Zhang, and Q.~Yang, ``Vertical federated learning,'' {\em arXiv preprint arXiv:2211.12814}, 2022.

\bibitem{Yang2023SurveyVFL}
L.~Yang, D.~Chai, J.~Zhang, Y.~Jin, L.~Wang, H.~Liu, H.~Tian, Q.~Xu, and K.~Chen, ``A survey on vertical federated learning: From a layered perspective,'' {\em arXiv preprint arXiv:2304.01829}, 2023.

\bibitem{Khan2022CommunicationEfficientVFL}
A.~Khan, M.~T. Thij, and A.~Wilbik, ``Communication-efficient vertical federated learning,'' {\em Algorithms}, vol.~15, no.~8, p.~273, 2022.

\bibitem{Serpanos2023MalwareDetectionVFL}
D.~Serpanos and G.~Xenos, ``Vertical federated learning in malware detection for smart cities,'' in {\em IEEE International Conference on Security}, 2023.

\bibitem{FuLabelInferenceVFL}
C.~Fu, X.~Zhang, S.~Ji, J.~Chen, J.~Wu, S.~Guo, J.~Zhou, A.~X. Liu, and T.~Wang, ``Label inference attacks against vertical federated learning,'' 2022.
\newblock DBLP conference proceedings.

\bibitem{Jacot2018NeuralTK}
A.~Jacot, F.~Gabriel, and C.~Hongler, ``Neural tangent kernel: Convergence and generalization in neural networks,'' {\em Proceedings of the National Academy of Sciences}, vol.~115, no.~34, pp.~E7665--E7671, 2018.

\bibitem{Reiss2012ActivityMonitoring}
A.~Reiss and D.~Stricker, ``Introducing a new benchmarked dataset for activity monitoring,'' in {\em 2012 16th International Symposium on Wearable Computers}, pp.~108--109, IEEE, 2012.

\bibitem{Sait2020COVID19XRay}
U.~Sait, K.~G. Lal, S.~Prajapati, R.~Bhaumik, T.~Kumar, S.~Sanjana, and K.~Bhalla, ``Curated dataset for covid-19 posterior-anterior chest radiography images (x-rays),'' 2020.

\bibitem{yang2021medmnist}
J.~Yang, R.~Shi, and B.~Ni, ``Medmnist classification decathlon: A lightweight automl benchmark for medical image analysis,'' in {\em 2021 IEEE 18th International Symposium on Biomedical Imaging (ISBI)}, pp.~191--195, IEEE, 2021.

\bibitem{Yang2021MedMNISTv2}
J.~Yang, R.~Shi, D.~Wei, Z.~Liu, L.~Zhao, B.~Ke, H.~Pfister, and B.~Ni, ``{MedMNIST v2-a large-scale lightweight benchmark for 2d and 3d biomedical image classification},'' {\em Scientific Data}, vol.~10, no.~1, p.~41, 2023.

\bibitem{Bilic2019LiTS}
P.~Bilic, P.~F. Christ, E.~Vorontsov, G.~Chlebus, H.~Chen, Q.~Dou, C.-W. Fu, X.~Han, P.-A. Heng, J.~Hesser, {\em et~al.}, ``The liver tumor segmentation benchmark (lits),'' {\em arXiv preprint arXiv:1901.04056}, 2019.

\bibitem{Xu2019OrganLocalization}
X.~Xu, F.~Zhou, B.~Liu, D.~Fu, and X.~Bai, ``Efficient multiple organ localization in ct image using 3d region proposal network,'' {\em IEEE Transactions on Medical Imaging}, vol.~38, no.~8, pp.~1885--1898, 2019.

\bibitem{hinton2015distilling}
G.~Hinton, O.~Vinyals, and J.~Dean, ``Distilling the knowledge in a neural network,'' {\em arXiv preprint arXiv:1503.02531}, 2015.

\bibitem{Yang2024KD}
D.~Yang {\em et~al.}, ``Federated learning with knowledge distillation for multi-organ segmentation with partially labeled datasets,'' {\em Medical Image Analysis}, vol.~95, p.~102967, 2024.

\bibitem{Lin2020EnsembleDistillation}
T.~Lin, L.~Kong, S.~U. Stich, and M.~Jaggi, ``Ensemble distillation for robust model fusion in federated learning,'' {\em arXiv preprint arXiv:2006.07242}, 2020.

\bibitem{Hsu2019NonIdenticalDistribution}
T.-M.~H. Hsu, H.~Qi, and M.~Brown, ``Measuring the effects of non-identical data distribution for federated visual classification,'' {\em arXiv preprint arXiv:1909.06335}, 2019.

\bibitem{khan2024brain}
H.~Khan, N.~C. Bouaynaya, and G.~Rasool, ``Brain-inspired continual learning: Robust feature distillation and re-consolidation for class incremental learning,'' {\em IEEE Access}, 2024.

\bibitem{Khan223Importance}
H.~Khan, N.~Bouaynaya, and G.~Rasool, ``{The Importance of Robust Features in Mitigating Catastrophic Forgetting},'' in {\em accepted for publication in 28th IEEE Symposium on Computers and Communications (ISCC 2023)}, 2023.
\newblock \url{https://arxiv.org/abs/2306.17091}.

\bibitem{Foley_2022}
{Patrick Foley and Micah J Sheller and Brandon Edwards and Sarthak Pati and Walter Riviera and Mansi Sharma and Prakash Narayana Moorthy and Shih-han Wang and Jason Martin and Parsa Mirhaji and Prashant Shah and Spyridon Bakas}, ``Openfl: the open federated learning library,'' {\em Physics in Medicine \& Biology}, vol.~67, p.~214001, oct 2022.

\bibitem{MONAI-Deploy}
{MONAI Team}, ``{MONAI Deploy},'' 2023.
\newblock available at: \url{https://monai.io/deploy.html}. Last accessed on Feb 1, 2023.

\bibitem{NVIDIA-CLARA}
{NVIDIA Team}, ``{NVIDIA Clara},'' 2023.
\newblock available at: \url{https://www.nvidia.com/en-us/clara/}. Last accessed on Feb 1, 2023.

\bibitem{Roth2022FLARE}
H.~Roth, Y.~Cheng, Y.~Wen, I.~Yang, Z.~Xu, Y.-T. Hsieh, K.~Kersten, A.~Harouni, C.~Zhao, K.~Lu, Z.~Zhang, W.~Li, A.~Myronenko, D.~Yang, S.~Yang, N.~Rieke, A.~Quraini, C.~Chen, D.~Xu, and A.~Feng, ``Nvidia flare: Federated learning from simulation to real-world,'' 10 2022.

\bibitem{huang2021ntk}
B.~Huang, X.~Li, Z.~Song, and X.~Yang, ``Fl-ntk: A neural tangent kernel-based framework for federated learning analysis,'' in {\em Proceedings of the 38th International Conference on Machine Learning} (M.~Meila and T.~Zhang, eds.), vol.~139 of {\em Proceedings of Machine Learning Research}, pp.~4423--4434, PMLR, 18--24 Jul 2021.

\bibitem{cifar10}
A.~Krizhevsky and G.~Hinton, ``Cifar-10 (canadian institute for advanced research),'' tech. rep., Canadian Institute for Advanced Research, 2009.

\bibitem{9069945}
K.~Wei, J.~Li, M.~Ding, C.~Ma, H.~H. Yang, F.~Farokhi, S.~Jin, T.~Q.~S. Quek, and H.~Vincent~Poor, ``Federated learning with differential privacy: Algorithms and performance analysis,'' {\em IEEE Transactions on Information Forensics and Security}, vol.~15, pp.~3454--3469, 2020.

\bibitem{nampalle2023vis}
K.~B. Nampalle, P.~Singh, U.~Narayan, and B.~Raman, ``Vision through the veil: Differential privacy in federated learning for medical image classification,'' 06 2023.

\bibitem{asad2020opt}
M.~Asad, A.~Moustafa, and T.~Ito, ``Fedopt: Towards communication efficiency and privacy preservation in federated learning,'' {\em Applied Sciences}, vol.~10, pp.~1--17, 04 2020.

\bibitem{TRUHN2023103059}
D.~Truhn, S.~T. Arasteh, O.~L. Saldanha, G.~Müller-Franzes, F.~Khader, P.~Quirke, N.~P. West, R.~Gray, G.~G. Hutchins, J.~A. James, M.~B. Loughrey, M.~Salto-Tellez, H.~Brenner, A.~Brobeil, T.~Yuan, J.~Chang-Claude, M.~Hoffmeister, S.~Foersch, T.~Han, S.~Keil, M.~Schulze-Hagen, P.~Isfort, P.~Bruners, G.~Kaissis, C.~Kuhl, S.~Nebelung, and J.~N. Kather, ``Encrypted federated learning for secure decentralized collaboration in cancer image analysis,'' {\em Medical Image Analysis}, p.~103059, 2023.

\bibitem{truex2019hybrid}
S.~Truex, N.~Baracaldo, A.~Anwar, T.~Steinke, H.~Ludwig, R.~Zhang, and Y.~Zhou, ``A hybrid approach to privacy-preserving federated learning,'' AISec'19, (New York, NY, USA), p.~1–11, Association for Computing Machinery, 2019.

\bibitem{Qi2023}
T.~Qi, F.~Wu, C.~Wu, L.~He, Y.~Huang, and X.~Xie, ``Differentially private knowledge transfer for federated learning,'' {\em Nature Communications}, vol.~14, p.~3785, June 2023.

\bibitem{So2023secure}
J.~So, R.~E. Ali, B.~G\"{u}ler, J.~Jiao, and A.~S. Avestimehr, ``Securing secure aggregation: Mitigating multi-round privacy leakage in federated learning,'' AAAI'23/IAAI'23/EAAI'23, AAAI Press, 2023.

\bibitem{wang2023LDS}
T.~Wang, Q.~Yang, K.~Zhu, J.~Wang, C.~Su, and K.~Sato, ``Lds-fl: Loss differential strategy based federated learning for privacy preserving,'' {\em IEEE Transactions on Information Forensics and Security}, vol.~19, pp.~1015--1030, 2024.

\bibitem{Xu2022DeTrustFLPF}
R.~Xu, N.~Baracaldo, Y.~Zhou, A.~Anwar, S.~Kadhe, and H.~Ludwig, ``Detrust-fl: Privacy-preserving federated learning in decentralized trust setting,'' {\em 2022 IEEE 15th International Conference on Cloud Computing (CLOUD)}, pp.~417--426, 2022.

\bibitem{LU2022102298}
M.~Y. Lu, R.~J. Chen, D.~Kong, J.~Lipkova, R.~Singh, D.~F. Williamson, T.~Y. Chen, and F.~Mahmood, ``Federated learning for computational pathology on gigapixel whole slide images,'' {\em Medical Image Analysis}, vol.~76, p.~102298, 2022.

\bibitem{li2019privacy}
W.~Li, F.~Milletar{\`i}, D.~Xu, N.~Rieke, J.~Hancox, W.~Zhu, M.~Baust, Y.~Cheng, S.~Ourselin, M.~Cardoso, and A.~Feng, ``Privacy-preserving federated brain tumour segmentation,'' {\em Hindawi}, 2019.

\bibitem{nanayakkara2022visualizing}
P.~Nanayakkara, J.~Bater, X.~He, J.~Hullman, and J.~Duggan, ``Visualizing privacy-utility trade-offs in differentially private data releases,'' {\em Proceedings on Privacy Enhancing Technologies}, 2022.

\bibitem{xu2021privacy}
L.~Xu, C.~Jiang, Y.~Qian, J.~Li, Y.~Zhao, and Y.~Ren, ``Privacy-accuracy trade-off in differentially-private distributed classification: A game theoretical approach,'' {\em IEEE Transactions on Big Data}, 2021.

\bibitem{Adnan2021}
M.~Adnan, S.~Kalra, J.~C. Cresswell, G.~W. Taylor, and H.~Tizhoosh, ``Federated learning and differential privacy for medical image analysis,'' {\em Nature}, 2021.

\bibitem{Choudhury2019DifferentialPF}
O.~Choudhury, A.~Gkoulalas-Divanis, T.~Salonidis, I.~Sylla, Y.~Park, G.~Hsu, and A.~K. Das, ``Differential privacy-enabled federated learning for sensitive health data,'' {\em ArXiv}, vol.~abs/1910.02578, 2019.

\bibitem{Dhiman2023}
S.~Dhiman, S.~Nayak, G.~K. Mahato, A.~Ram, and S.~K. Chakraborty, ``Homomorphic encryption based federated learning for financial data security,'' in {\em IEEE International Conference on Innovations in Computer Science and Engineering (I3CS)}, 2023.

\bibitem{Stripelis2021}
D.~Stripelis, H.~Saleem, T.~Ghai, N.~J. Dhinagar, U.~Gupta, C.~Anastasiou, G.~V. Steeg, S.~Ravi, M.~Naveed, P.~M. Thompson, and J.~Ambite, ``Secure neuroimaging analysis using federated learning with homomorphic encryption,'' in {\em SPIE Medical Imaging}, 2021.

\bibitem{Burlachenko2023}
K.~Burlachenko, A.~Alrowithi, F.~A. Albalawi, and P.~Richtárik, ``Federated learning is better with non-homomorphic encryption,'' in {\em Proceedings of the ACM Symposium on Cloud Computing}, 2023.

\bibitem{acar2017survey}
A.~Acar, H.~Aksu, A.~S. Uluagac, and M.~Conti, ``A survey on homomorphic encryption schemes: Theory and implementation,'' 2017.

\bibitem{menze2014multimodal}
B.~H. Menze, A.~Jakab, S.~Bauer, J.~Kalpathy-Cramer, K.~Farahani, J.~Kirby, Y.~Burren, N.~Porz, J.~Slotboom, R.~Wiest, {\em et~al.}, ``The multimodal brain tumor image segmentation benchmark (brats),'' {\em IEEE transactions on medical imaging}, vol.~34, no.~10, pp.~1993--2024, 2015.

\bibitem{TCGACOADREAD}
T.~C. G.~A. Network, ``Comprehensive molecular characterization of human colon and rectal cancer,'' {\em Nature}, vol.~487, no.~7407, pp.~330--337, 2012.

\bibitem{lecun1998gradient}
Y.~LeCun, L.~Bottou, Y.~Bengio, and P.~Haffner, ``Gradient-based learning applied to document recognition,'' {\em Proceedings of the IEEE}, vol.~86, no.~11, pp.~2278--2324, 1998.

\bibitem{Ryu2022appfl}
M.~Ryu, Y.~Kim, K.~Kim, and R.~K. Madduri, ``Appfl: Open-source software framework for privacy-preserving federated learning,'' in {\em 2022 IEEE International Parallel and Distributed Processing Symposium Workshops (IPDPSW)}, (Los Alamitos, CA, USA), pp.~1074--1083, IEEE Computer Society, jun 2022.

\bibitem{boyd2011distributed}
S.~Boyd, N.~Parikh, E.~Chu, B.~Peleato, and J.~Eckstein, ``Distributed optimization and statistical learning via the alternating direction method of multipliers,'' {\em Foundations and Trends in Machine Learning}, vol.~3, no.~1, pp.~1--122, 2011.

\bibitem{KaissisEnd}
G.~Kaissis, A.~Ziller, J.~Passerat-Palmbach, T.~Ryffel, D.~Usynin, A.~Trask, I.~Lima, J.~Mancuso, F.~Jungmann, M.-M. Steinborn, A.~Saleh, M.~Makowski, D.~Rueckert, and R.~Braren, ``End-to-end privacy preserving deep learning on multi-institutional medical imaging,'' {\em Nature Machine Intelligence}, vol.~3, pp.~1--12, 06 2021.

\bibitem{Shi2023fedensemble}
N.~Shi, F.~Lai, R.~A. Kontar, and M.~Chowdhury, ``Fed-ensemble: Ensemble models in federated learning for improved generalization and uncertainty quantification,'' {\em IEEE Transactions on Automation Science and Engineering}, pp.~1--0, 2023.

\bibitem{plassier_conformal}
V.~Plassier, M.~Makni, A.~Rubashevskii, E.~Moulines, and M.~Panov, ``Conformal prediction for federated uncertainty quantification under label shift,'' in {\em SEFM}, vol.~11724 of {\em Lecture Notes in Computer Science}, pp.~183--202, Springer, 2023.

\bibitem{Lu2023conformal}
C.~Lu, Y.~Yu, S.~P. Karimireddy, M.~I. Jordan, and R.~Raskar, ``Federated conformal predictors for distributed uncertainty quantification,'' in {\em Proceedings of the 40th International Conference on Machine Learning}, ICML'23, JMLR.org, 2023.

\bibitem{bhatt2023federated}
S.~Bhatt, A.~Gupta, and P.~Rai, ``Federated learning with uncertainty via distilled predictive distributions,'' 2023.

\bibitem{FedUA2023}
S.-M. Lee and J.-L. Wu, ``Fedua: An uncertainty-aware distillation-based federated learning scheme for image classification,'' {\em Information}, vol.~14, p.~234, Apr 2023.

\bibitem{makhija2023privacy}
D.~Makhija, J.~Ghosh, and N.~Ho, ``Privacy preserving bayesian federated learning in heterogeneous settings,'' {\em arXiv preprint arXiv:2306.07959}, 2023.

\bibitem{zhang2023uncertainty}
Y.~Zhang, T.~Xia, A.~Ghosh, and C.~Mascolo, ``Uncertainty quantification in federated learning for heterogeneous health data,'' in {\em International Workshop on Federated Learning for Distributed Data Mining}, 2023.

\bibitem{Chen2022SelfAwarePF}
H.~Chen, J.~Ding, E.~W. Tramel, S.~Wu, A.~K. Sahu, S.~Avestimehr, and T.~Zhang, ``Self-aware personalized federated learning,'' {\em ArXiv}, vol.~abs/2204.08069, 2022.

\bibitem{zhang2022vari}
X.~Zhang, Y.~Li, W.~Li, K.~Guo, and Y.~Shao, ``Personalized federated learning via variational {B}ayesian inference,'' in {\em Proceedings of the 39th International Conference on Machine Learning} (K.~Chaudhuri, S.~Jegelka, L.~Song, C.~Szepesvari, G.~Niu, and S.~Sabato, eds.), vol.~162 of {\em Proceedings of Machine Learning Research}, pp.~26293--26310, PMLR, 17--23 Jul 2022.

\bibitem{kotelevskii2022pop}
N.~Kotelevskii, M.~Vono, A.~Durmus, and E.~Moulines, ``Fedpop: A bayesian approach for personalised federated learning,'' in {\em Advances in Neural Information Processing Systems} (S.~Koyejo, S.~Mohamed, A.~Agarwal, D.~Belgrave, K.~Cho, and A.~Oh, eds.), vol.~35, pp.~8687--8701, Curran Associates, Inc., 2022.

\bibitem{zhou2023FA}
T.~Zhou, J.~Zhang, and D.~H. Tsang, ``Fedfa: Federated learning with feature anchors to align features and classifiers for heterogeneous data,'' {\em IEEE Transactions on Mobile Computing}, pp.~1--12, 2023.

\bibitem{Thorgeirsson2020ProbabilisticPW}
A.~T. Thorgeirsson and F.~Gauterin, ``Probabilistic predictions with federated learning,'' {\em Entropy}, vol.~23, 2020.

\bibitem{Luo2021NoFO}
M.~Luo, F.~Chen, D.~Hu, Y.~Zhang, J.~Liang, and J.~Feng, ``No fear of heterogeneity: Classifier calibration for federated learning with non-iid data,'' {\em ArXiv}, vol.~abs/2106.05001, 2021.

\bibitem{QI23silo}
Z.~Qi, L.~Meng, Z.~Chen, H.~Hu, H.~Lin, and X.~Meng, ``Cross-silo prototypical calibration for federated learning with non-iid data,'' MM '23, (New York, NY, USA), p.~3099–3107, Association for Computing Machinery, 2023.

\bibitem{sanderson2018uncertainty}
B.~Sanderson, ``Uncertainty quantification in multi-model ensembles,'' {\em Oxford Research Encyclopedia of Climate Science}, 2018.

\bibitem{CIFAR}
A.~Krizhevsky, ``Learning multiple layers of features from tiny images. university of toronto (2012),'' {\em \url{http://www.cs.toronto.edu/kriz/cifar.html}, last accessed}, vol.~5, p.~13, 2022.

\bibitem{Kuznetsova2020OpenImagesV4}
A.~Kuznetsova, H.~Rom, N.~Alldrin, J.~Uijlings, I.~Krasin, J.~Pont-Tuset, S.~Kamali, S.~Popov, M.~Malloci, A.~Kolesnikov, {\em et~al.}, ``The open images dataset v4,'' {\em International Journal of Computer Vision}, pp.~1--26, 2020.

\bibitem{gammerman98conf}
A.~Gammerman, V.~Vovk, and V.~Vapnik, ``Learning by transduction,'' in {\em Proceedings of the Fourteenth Conference on Uncertainty in Artificial Intelligence}, UAI'98, (San Francisco, CA, USA), p.~148–155, Morgan Kaufmann Publishers Inc., 1998.

\bibitem{saunders99conf}
C.~Saunders, A.~Gammerman, and V.~Vovk, ``Transduction with confidence and credibility.,'' pp.~722--726, 01 1999.

\bibitem{Lu2021DistributionFreeFL}
C.~Lu and J.~Kalpathy-Cramer, ``Distribution-free federated learning with conformal predictions,'' {\em ArXiv}, vol.~abs/2110.07661, 2021.

\bibitem{ImageNet}
O.~Russakovsky, J.~Deng, H.~Su, J.~Krause, S.~Satheesh, S.~Ma, Z.~Huang, A.~Karpathy, A.~Khosla, M.~S. Bernstein, A.~C. Berg, and L.~Fei{-}Fei, ``Imagenet large scale visual recognition challenge,'' {\em Int. J. Comput. Vis.}, vol.~115, no.~3, pp.~211--252, 2015.

\bibitem{vovk2005algorithmic}
V.~Vovk, A.~Gammerman, and G.~Shafer, {\em Algorithmic Learning in a Random World}.
\newblock Statistics for Engineering and Information Science, Springer, 2005.

\bibitem{brown2022simple}
K.~E. Brown and D.~A. Talbert, ``A simple direct uncertainty quantification technique based on machine learning regression,'' {\em FLAIRS Conference}, 2022.

\bibitem{swaminathan2022bayesian}
M.~Swaminathan, O.~W. Bhatti, Y.~Guo, E.~Huang, and O.~Akinwande, ``Bayesian learning for uncertainty quantification, optimization, and inverse design,'' {\em IEEE Transactions on Microwave Theory and Techniques}, 2022.

\bibitem{DALMAZ2024103121}
O.~Dalmaz, M.~U. Mirza, G.~Elmas, M.~Ozbey, S.~U. Dar, E.~Ceyani, K.~K. Oguz, S.~Avestimehr, and T.~Çukur, ``One model to unite them all: Personalized federated learning of multi-contrast mri synthesis,'' {\em Medical Image Analysis}, vol.~94, p.~103121, 2024.

\bibitem{femnist}
S.~Caldas, S.~M.~K. Duddu, P.~Wu, T.~Li, J.~Kone{\v{c}}n{\`y}, H.~B. McMahan, V.~Smith, and A.~Talwalkar, ``Federated extended mnist (femnist).'' \url{https://github.com/TalwalkarLab/leaf}, 2018.

\bibitem{sent140}
A.~Go, R.~Bhayani, and L.~Huang, ``Sentiment140.'' \url{http://help.sentiment140.com/for-students}, 2009.

\bibitem{breiman1996bagging}
L.~Breiman, ``Bagging predictors,'' {\em Machine Learning}, vol.~24, no.~2, pp.~123--140, 1996.

\bibitem{pearce2021understanding}
T.~Pearce, A.~Brintrup, and J.~Zhu, ``Understanding softmax confidence and uncertainty,'' {\em arXiv preprint arXiv:2106.04972}, 2021.

\bibitem{tinyimagenet}
``Tiny imagenet.'' \url{https://tiny-imagenet.herokuapp.com/}.

\bibitem{vireofood172}
``Vireofood-172.'' \url{https://fvl.fudan.edu.cn/dataset/vireofood172/list.htm}.

\bibitem{pati2021federated}
S.~Pati, U.~Baid, M.~Zenk, B.~Edwards, M.~Sheller, G.~A. Reina, P.~Foley, A.~Gruzdev, J.~Martin, S.~Albarqouni, Y.~Chen, R.~T. Shinohara, A.~Reinke, D.~Zimmerer, J.~B. Freymann, J.~S. Kirby, C.~Davatzikos, R.~R. Colen, A.~Kotrotsou, D.~Marcus, M.~Milchenko, A.~Nazeri, H.~Fathallah-Shaykh, R.~Wiest, A.~Jakab, M.-A. Weber, A.~Mahajan, L.~Maier-Hein, J.~Kleesiek, B.~Menze, K.~Maier-Hein, and S.~Bakas, ``The federated tumor segmentation (fets) challenge,'' 2021.

\bibitem{ronneberger2015unet}
O.~Ronneberger, P.~Fischer, and T.~Brox, ``U-net: Convolutional networks for biomedical image segmentation,'' in {\em Medical Image Computing and Computer-Assisted Intervention – MICCAI 2015}, (Cham), pp.~234--241, Springer International Publishing, 2015.

\bibitem{pati2022federated}
S.~Pati, U.~Baid, B.~Edwards, M.~Sheller, S.-H. Wang, G.~A. Reina, P.~Foley, A.~Gruzdev, D.~Karkada, C.~Davatzikos, {\em et~al.}, ``Federated learning enables big data for rare cancer boundary detection,'' {\em Nature communications}, vol.~13, no.~1, p.~7346, 2022.

\end{thebibliography}
\bibliographystyle{ieeetr}

\end{document}